\documentclass[runningheads]{llncs}

\usepackage{eccv}

\usepackage{eccvabbrv}

\usepackage{graphicx}
\usepackage{booktabs}

\usepackage[accsupp]{axessibility}  

\usepackage{multirow}
\usepackage{makecell}
\usepackage{caption}
\usepackage{arydshln}
\usepackage{subcaption}
\usepackage{array}
\usepackage{amssymb}
\usepackage{pifont}

\newcommand{\phseo}[1]{\textcolor{blue}{{[Paul: #1]}}}

\usepackage[pagebackref,breaklinks,colorlinks,citecolor=eccvblue]{hyperref}

\usepackage{orcidlink}

\newcommand{\sh}[1]{\textcolor{red}{[SHYu: #1]}}

\begin{document}


\title{Pseudo-RIS: Distinctive Pseudo-supervision Generation for Referring Image Segmentation} 


\titlerunning{Pseudo-RIS}

\author{Seonghoon Yu\inst{1} 
\quad 
Paul Hongsuck Seo\inst{2}$^\dagger$ 
\quad 
Jeany Son\inst{1}$^\dagger$
}

\authorrunning{Yu et al.}

\institute{\hspace{-0.15cm}AI Graduate School, GIST, Korea \and
\hspace{-0.1cm}Department of Computer Science and Engineering, Korea University, Korea
\\
\email{seonghoon@gm.gist.ac.kr \quad phseo@korea.ac.kr \quad jeany@gist.ac.kr}
\\
\url{https://github.com/Seonghoon-Yu/Pseudo-RIS} 
}


\maketitle

\begin{abstract}

We propose a new framework that automatically generates high-quality segmentation masks with their referring expressions as pseudo supervisions for referring image segmentation (RIS).
These pseudo supervisions allow the training of any supervised RIS methods without the cost of manual labeling.
To achieve this, we incorporate existing segmentation and image captioning foundation models, leveraging their broad generalization capabilities.
However, the na\"ive incorporation of these models may generate non-distinctive expressions that do not distinctively refer to the target masks.
To address this challenge, we propose two-fold strategies that generate distinctive captions:
1) `distinctive caption sampling', a new decoding method for the captioning model, to generate multiple expression candidates with detailed words focusing on the target.
2) `distinctiveness-based text filtering' to further validate the candidates and filter out those with a low level of distinctiveness.
These two strategies ensure that the generated text supervisions can distinguish the target from other objects, making them appropriate for the RIS annotations.
Our method significantly outperforms both weakly and zero-shot SoTA methods on the RIS benchmark datasets.
It also surpasses fully supervised methods in unseen domains, proving its capability to tackle the open-world challenge within RIS. 
Furthermore, integrating our method with human annotations yields further improvements, highlighting its potential in semi-supervised learning applications.
\renewcommand{\thefootnote}{}
\footnotetext{$^\dagger$Corresponding authors.}
\renewcommand{\thefootnote}{\arabic{footnote}}

\end{abstract}



\section{Introduction}
\label{sec:intro}

Referring image segmentation (RIS) task aims to segment the object referred by a given natural language description and has substantial applications across various fields such as human-object interaction~\cite{interaction}, image editing~\cite{image_editing}.
While RIS is particularly challenging because it requires the model to map and understand both visual data and textual descriptions simultaneously, recent fully supervised approaches~\cite{etris,cris, restr, vlt, fully_ris_1, fully_ris_2} have shown remarkable advances.
However, these methods heavily rely on costly human annotated datasets which include pixel-wise instance masks with their corresponding referring descriptions.
Moreover, the models trained on heavily supervised datasets show limited generalization performances~\cite{global_local_clip}.
This issue is particularly amplified by the current mainstream RIS datasets that mainly focus only on a few object classes (\eg, 80 classes in the RefCOCO variants).


\begin{figure}[t]
     \centering
     \begin{subfigure}[b]{0.31\textwidth}
         \centering
         \includegraphics[width=\textwidth]{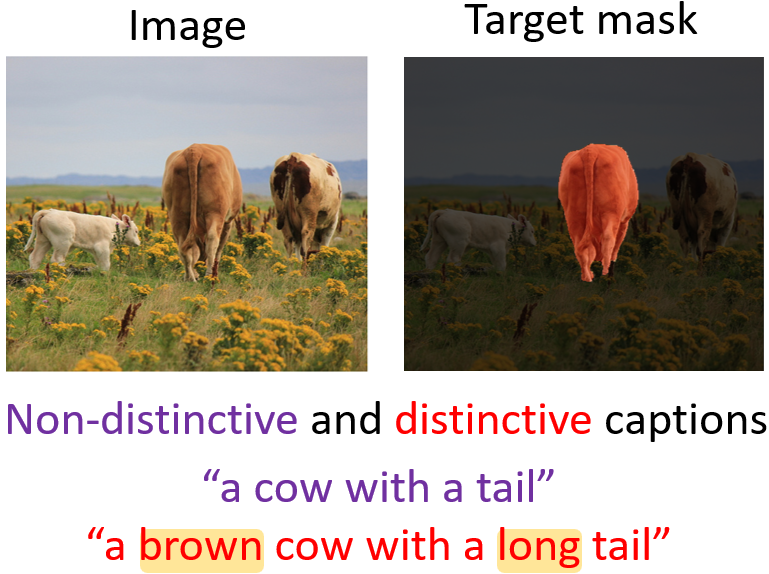}
         \caption{Caption examples in RIS}
         \label{fig:intro_a}
     \end{subfigure}
     \hfill
     \begin{subfigure}[b]{0.65\textwidth}
         \centering
         \includegraphics[width=\textwidth]{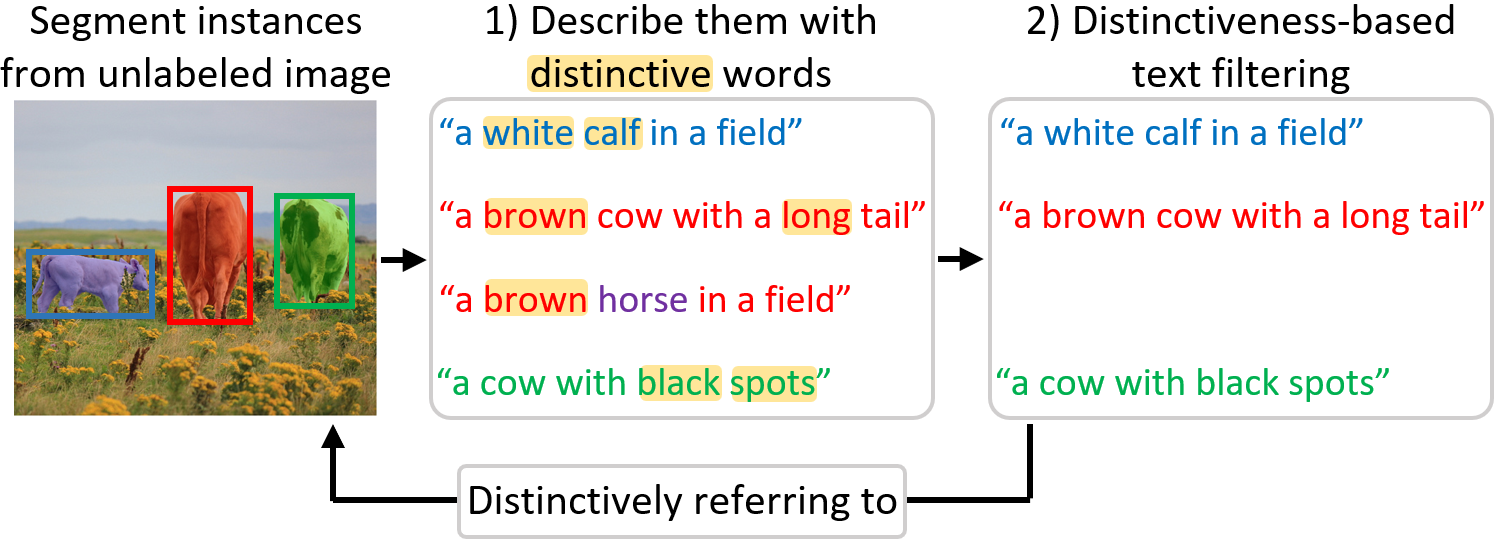}
         \caption{Distinctive Caption Generation}
         \label{fig:intro_b}
     \end{subfigure}
        \caption{Illustration of our distinctive pseudo-supervision generation: (a) a distinctive caption \textit{``a brown cow with a long tail"} distinctively refers to a target mask, while a non-distinctive caption \textit{``a cow with a tail"} causes misleading to a non-targeted object since there is another cow with a tail in an image, (b) two proposed methods for distinctive caption generation given segmentation masks: 1) multiple caption candidates generation with target-specific words; 2) filtering out the misleading captions among all candidates.}
        \label{fig:intro}
\end{figure}

A couple of recent works have proposed weakly supervised~\cite{tris, shatter_and_gather, tseg, weakly_ris_1} and zero-shot RIS approaches~\cite{global_local_clip, ref_diff} to handle costly human annotation.
On one hand, weakly supervised methods attempt to tackle RIS using the image-text paired data where the spatial groundings of language descriptions are missing.
On the other hand, zero-shot RIS approaches utilize large-scale vision-language pre-trained models~\cite{clip, stable_diffusion} and use their discriminative power to localize language expressions within an input image.
However, both the weakly supervised RIS models and the pre-trained models for the zero-shot methods are trained on the image-text paired datasets which lack grounding annotations, and furthermore the pre-trained models use scene-level captions that are often ambiguous to refer to a specific object.
Therefore, the final models have limitations in precisely localizing references. 

In this paper, we resolve these issues by introducing a novel framework where high-quality segmentation masks together with target-specific distinctive referring expressions are automatically generated.
We achieve this by leveraging multiple foundation models with strong generalization powers.
The automatically generated data then serve as pseudo RIS annotations that can train any supervised RIS methods at scale.
Moreover, the broad domain coverage of the foundation models makes our pseudo-supervisions appropriate for the open-world nature of RIS.

In particular, we incorporate a segmentation foundation model, namely SAM\cite{sam}, and a text-generative foundation model such as CoCa~\cite{coca}.
The former extracts precise class-agnostic segmentation masks; the latter then generates a target-specific caption given the image region obtained from one of the generated masks as the input.
However, although this naive approach may generate a valid caption for the input image region, the generated caption may not distinctively refer to the target object.
For example, Fig.~\ref{fig:intro_a} shows a correct but non-distinctive caption in \textcolor{violet}{purple} for the input image region of a generated instance mask.
The caption \textit{``a cow with a tail"} correctly describes the content of the input region but does not distinctively refer to the target region; there is another cow with a tail, making the generated caption not appropriate as a referring expression for the input region. 
In contrast, the other caption \textit{``a brown cow with a long tail"} in \textcolor{red}{red} distinctively refers to the target region of the single cow.

To generate such a distinctive caption, we propose two techniques that allow distinctive caption generation, as in Fig.~\ref{fig:intro_b}: 
1) \textit{distinctive caption sampling} for generating candidate referring expressions with details in the target region, and 2) \textit{distinctiveness-based text filtering} that further validates distinctiveness of generated captions.
These two techniques enable the final captions to not only accurately describe the target, but also distinguish it from other objects within the same image, making the generated caption appropriate for RIS training.
Furthermore, our method allows to naturally integrate the generalization powers of the strong foundation models into our generated dataset and consequently, the trained models on our framework achieve superior performances in various challenging RIS configurations.

Our contributions can be summarised as follows:
\begin{itemize}
    \item {We introduce a novel framework of automatically generating pseudo supervision for RIS that allows to train any supervised RIS methods with no human labeling cost.}

    \item {By incorporating with foundation models, our method not only generates precise pseudo supervisions for RIS but also successfully addresses generalization problems in supervised RIS. }

    \item {We propose \textit{distinctive caption sampling} for generating caption candidates with target-specific expressions and \textit{distinctiveness-based text filtering} for further validation of the generated referring expressions.}

    \item {Our extensive set of experiments demonstrates that trained models under our framework achieve state-of-the-art performances in various RIS configurations and show strong generalization performances in grounding unseen objects.}
    

\end{itemize}

\section{Related Work}
\label{sec:related}

\subsection{Pseudo Supervision Generation.}
Several works have explored generating pseudo supervision~\cite{recent_pseudo_1,recent_pseudo_2, recent_pseudo_3, recent_pseudo_4,recent_pseudo_6} without human labels to address expensive annotation costs in tasks like natural language video localization~\cite{ps_1_nlvl}, video sentence localization~\cite{ps_3_video_sentence_localization}, video corpus moment retrieval~\cite{ps_4_video_moment_retrieval}, and referring expression comprehension~\cite{ps_2_pseudo_q}.
In video-related tasks~\cite{ps_1_nlvl,ps_3_video_sentence_localization,ps_4_video_moment_retrieval}, pseudo temporal event boundaries are generated within videos, followed by pseudo language queries using captioning models~\cite{blip}. 
In referring expression comprehension task, Pseudo-Q~\cite{ps_2_pseudo_q} extract object bounding boxes with their category and attributes utilizing pre-trained detector~\cite{pre_traied_detector} on Visual Genorm dataset~\cite{visual_genome}, then they use heuristic algorithms to produce pseudo language queries based on the template comprising class name, attribute, and relationships.
Unlike Pseudo-Q~\cite{ps_2_pseudo_q}, instead of heuristic methods, we automatically generate pseudo language queries using an image captioning model~\cite{coca} combined with our proposed decoding strategy.




\subsection{Decoding Strategy.}
Decoding strategy~\cite{r_de_1,r_de_2,r_de_3,r_de_4,r_de_5,r_de_6,r_de_7} is a vital in auto-regressive language modeling like GPT~\cite{gpt2, gpt3} for progressing from previously generated words to the next work.
The decoding strategies can be grouped into search~\cite{contrastive_decoding, contrastive_decoding_2, contrastive_decoding_3} or sampling approaches~\cite{top-p, mirostat_decoding, locally_decoding}.
In search-based approaches, beam search maximizes the multiple of the probabilities of each word in a sequence of words.
Contrastive decoding~\cite{contrastive_decoding} maximizes the word probability difference between larger and smaller language models.
In sampling-based approaches, top-$k$ and top-$p$~\cite{top-p} sampling sample words from the fixed top-$k$ or top-$p$ word probability distribution.
Mirostate sampling~\cite{mirostat_decoding} uses the statistical information of previously generated words to chose the next word.
Our proposed strategy utilizes the word distributions of other images to sample the next word focused on a target image.





\subsection{Referring Image Segmentation.}
Fully supervised RIS approaches~\cite{lavt, restr, vlt, efn, etris, cris, fully_ris_1, fully_ris_2, fully_ris_3, fully_ris_4, fully_ris_5, fully_ris_6, fully_ris_7, fully_ris_8, fully_ris_9, fully_ris_10, fully_ris_11, fully_ris_12, fully_ris_13, fully_ris_14, fully_ris_15, fully_ris_16, a_ris_1, a_ris_10, a_ris_11, a_ris_12, a_ris_14, a_ris_2, a_ris_3, a_ris_4, a_ris_5, a_ris_6, a_ris_7, a_ris_8, a_ris_9,a_ris_15} have achieved remarkable performance by effectively fusing two different modalities.
However, they highly depend on expensive human annotated data.
To handle this, weakly and zero-shot RIS methods have emerged as potential solutions, recently.
Weakly supervised methods~\cite{shatter_and_gather, tris, tseg, weakly_ris_1} use only image-text pairs for training.
They mainly deal with overcoming the lack of localization for referred instances.
\textit{Lee et al.}~\cite{weakly_ris_1} utilize Grad-CAM~\cite{grad_cam} for localization, and SaG~\cite{shatter_and_gather} exploits slot attention~\cite{slot} to capture individual entities.
Zero-shot RIS methods~\cite{global_local_clip, ref_diff, tas} solve the RIS task without any additional training by leveraging the vision-language foundation models~\cite{clip, stable_diffusion} with the instance segmentation model~\cite{sam, freesolo}.
Unlike current weakly supervised and zero-shot RIS methods, we generate precise pseudo-supervisions for RIS to handle costly human annotation.



\section{Method}
\label{sec:method}



In this section, we present Pseudo-RIS, the novel framework for RIS that automatically generates distinctive referring expressions paired with their corresponding target segmentation masks.
We note that getting a distinctive referring text for each target region is critical particularly in RIS, as the resulting mask should change when a non-distinctive caption is given. For instance, in Fig.~\ref{fig:intro}(a), the mask for the non-distinctive caption should change to cover all cows that have a tail.
The proposed Pseudo-RIS consists of three steps: 1) \textit{precise mask extraction}, 2) \textit{distinctive caption candidate generation}, and 3) \textit{distinctiveness-based text filtering}.
In the mask generation step, we first obtain precise instance-level segmentation mask using a segmentation foundation model such as SAM~\cite{sam}.
Then, a text-generative foundation model, \ie, large-scale captioning model~\cite{coca}, is used to sample an extensive set of candidate referring expressions for each of the generated masks using our proposed \textit{distinctive caption sampling} method.
Finally, we further evaluate the captions generated for the given masks using our \textit{distinctiveness metric} derived from CLIP~\cite{clip} scores, and filter out ambiguous and imprecise captions~\cite{distinctiveness}.
In the following sections, we further elaborate on each step in more detail.

\subsection{Mask Extraction}
\label{subsec:mask_generation}
We first extract segmentation masks from an unlabeled image by exploiting a segmentation foundation model, SAM~\cite{sam}.
It allows us to obtain hiqh-quality and class-agnostic masks in the wild, leading to valuable pseudo-supervisions appropriate for the open-world nature of RIS.

\begin{figure*}[t]
    \centering
    \includegraphics[width=1.0\linewidth]{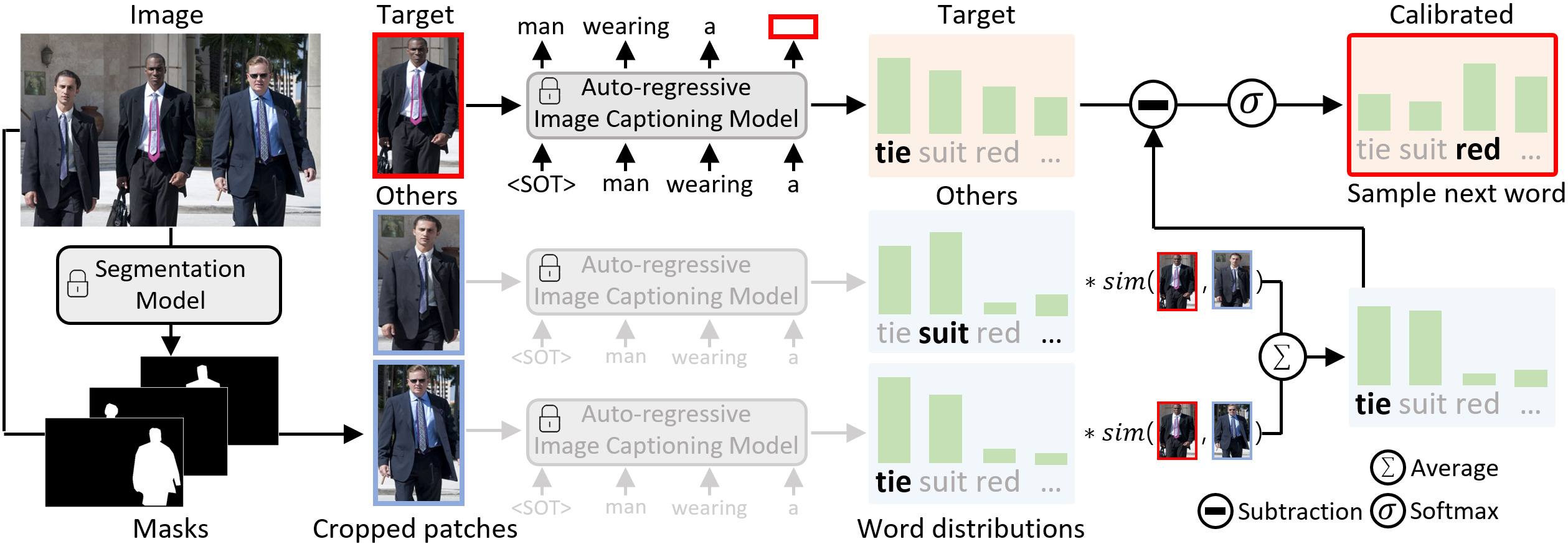}
    \caption{Distinctive caption candidates generation of our Pseudo-RIS. Given segmentation masks, we generate multiple distinctive caption candidates on each mask using a frozen image captioning model with the proposed \textit{distinctive caption sampling}. This method calibrates the word distribution of the target by utilizing that of others, and then samples the next word from a calibrated distribution of the target.} 
    \label{fig:ds}
\end{figure*}

\subsection{Distinctive Caption Candidates Generation}
\label{subsec:distinctive}

Given extracted segmentation masks within an image, we employ an image-text foundation model, such as CoCa~\cite{coca}, to generate multiple distinctive caption candidates for each mask.
To generate distinctive captions, we propose a new decoding strategy, termed \textit{distinctive caption sampling} based on the captioning model.
In the proposed decoding strategy, the captioning model is encouraged to produce target-specific words capturing dicriminative features of the target image region, which allows the generated captions to distinctively refer only to the target region.
For example, among several people as in Fig.~\ref{fig:ds}, a target is uniquey identified by the description \textit{``man wearing a red tie"} due to the target-specific word \textit{``red"}. 
In contrast, more generic words like \textit{``tie"} or \textit{``suit"} do not provide the same level of distinctiveness.
In this section, we demonstrate how to generate such distinctive captions using the frozen captioning model with our distinctive caption sampling.

\paragraph{Generating Caption for a Target Mask.}

Given a cropped patch for a given target mask as visual context, the image captioning model generates a sequence of $m$ words   $\mathbf{y} = \langle y_1, y_2, \dots, y_m \rangle$ from the vocabulary $\mathcal{V}$ ($y\in\mathcal{V}$), that form the description representing the image region of the target mask.
Consequently, the frozen auto-regressive captioning model computes a joint probability distribution of the sequence of words, $P(\mathbf{y}|x_i)$, in a left-to-right and token-by-token manner:
\begin{align}
        P(\textbf{y}|x_i) = \prod_{t=1}^{m}p(y_t|y_{<t}, x_i, C),
\end{align}
where $x_i$ is a cropped patch for a target $i$ with margins to capture surrounding contexts and $C$ denotes a set of all cropped patches for all extracted masks in a given image.
Each word in a description is generated by a specific decoding strategy, such as beam search, top-$k$, or top-$p$~\cite{top-p} sampling, which determines how the model progresses from the previously generated sequence of words $y_{< t}$ = $\langle y_1, \dots, y_{t-1} \rangle$ to the next word $y_{t}$ at each $t$ step.
We use different values of $k$ and $p$, along with different cropping margins for the patches,  to generate diverse multiple caption candidates.
For further details about these decoding strategies, which are commonly used in text generation tasks~\cite{top-k, top-p}, please refer to our supplementary material.

\paragraph{Distinctive Caption Sampling.}
\label{para:distinctive_caption_sampling}
This aims to sample the next word that is more specific to a target mask at each step.
To this end, we calibrate the probabilities of the words for a target patch by penalizing high-probability words in other patches and re-warding low-probability words in other patches.
In detail, for a given target cropped patch $x_i$, we calculate the probability distribution of the next word $P(y_{t}|y_{< t}, x_i)$, which is obtained by applying the softmax function to model output lozit $z_t$ (unnormalized score) at $t$ step. Simultaneously, we also compute this probability for all other cropped patches as $P(y_{t}|y_{< t}, x_j)$, where $x_j \in C \backslash \{x_i\}$. Notice that $y_{< t}$ is previously generated words for a target patch $x_i$.
Then, the calibrated word probability at $t$ step is given by:
\begin{align}
    &P(y_{t}|y_{< t}, x_i, C)\nonumber\\
    &= \sigma \big( P(y_{t}|y_{< t}, x_i) - \frac{1}{n} \hspace{-0.2cm} \sum_{ x_j \in C \backslash \{x_i\}} \hspace{-0.2cm} s_{ij}P(y_{t}|y_{< t}, x_j)  \big),
\end{align}
where $\sigma(\cdot)$ is the softmax function with a temperature, $n=|C|-1$ denotes the number of the extracted masks except a target $i$ in a given image, and $s_{ij}$ is a cosine similarity between visual embeddings of $x_i$ and $x_j$, extracted from the visual encoder of the image captioning model.
$s_{ij}$ modulates the influence of other patches, increasing it for patches similar to a target patch $x_i$, and decreasing it for those not similar, thereby aiding in sampling words that distinguish similar objects better. 
We sample the next word ${y}_t$ from this calibrated distribution:
\begin{align}
    {y}_{t} \sim P(y_{t}|{y}_{< t}, x_i, C).
\end{align}
The detailed illustration of the proposed distinctive caption sampling is shown in Fig.~\ref{fig:ds}.
In the original word distribution of a target, the generic words (\textit{``tie"} and \textit{''suit"}) have a high probability, but after calibrating the word distribution of a target, generic words are suppressed, and target-focused word \textit{''red"} has the highest probability, which is more likely to be selected for the next word $y_t$.

\subsection{Distinctiveness-based Text Filtering}
\label{subsec:filtering}
\begin{figure*}[t]
    \centering
    \includegraphics[width=0.98\linewidth]{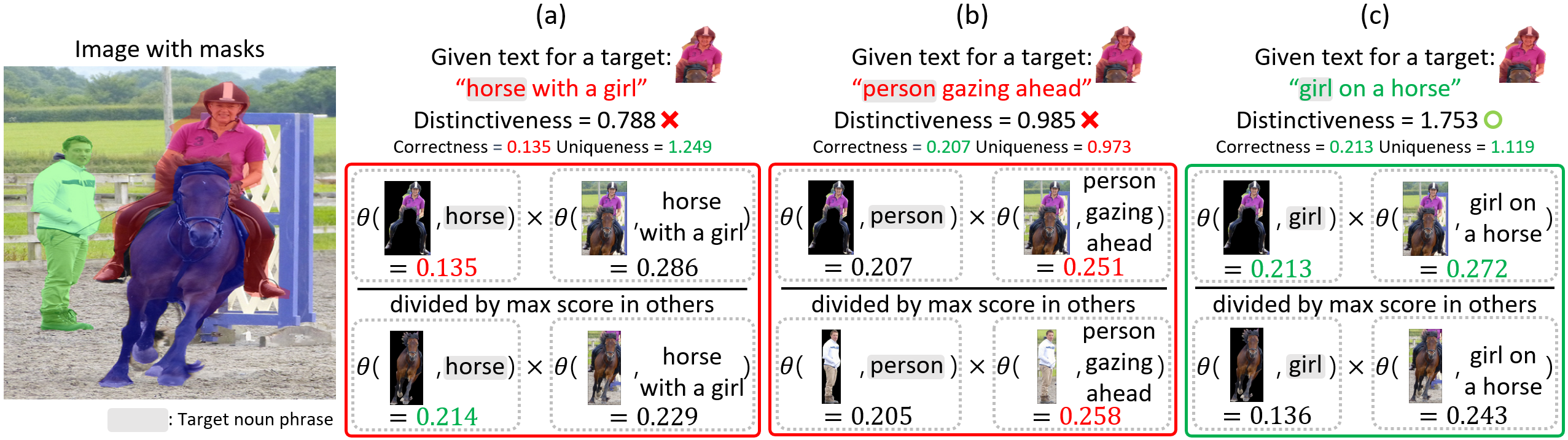}
    \caption{Distinctiveness-based text filtering of our Pseudo-RIS. On each mask, we filter out non-distinctive caption candidates with a \textit{distinctiveness score} below a threshold $\tau$. This distinctiveness score measures how well a caption distinctively refers to a target mask, with examples illustrating: (a) incorrect caption describing a non-targeted horse, resulting in a low correctness despite its high uniqueness, (b) ambiguous caption referring to an unintended mask, yielding low uniqueness but high correctness, and (c) a target-specific distinctive caption, achieving high uniqueness and correctness scores, leading to high distinctiveness.}
    \label{fig:distinctiveness}
\end{figure*}

In this section, we present our \textit{distinctiveness-based text filtering} method, that filter out ambiguous and imprecise captions given all the generated caption candidates for each mask, by utilizing the discriminative power of a foundation model, CLIP~\cite{clip}, to validate whether the captions correctly describes the target masks. 
The generated caption candidates may sometimes ambiguously refer to the unintended object or even incorrectly describe a non-targeted mask located around a target mask within a cropped patch.
This is because, (1) we employ a captioning model trained only on noisy non-curated web datasets, causing noisy captions, (2) a cropped patch for a target could contain other masks, (\eg a horse appears in a cropped patch for a girl, as in Fig.~\ref{fig:distinctiveness}, thus a captioning model could describe these non-targeted masks.
To tackle these issues, we introduce a \textit{distinctiveness score} that evaluates how well the generated captions \textbf{uniquely} and \textbf{correctly} identify the target mask.
Using this score, misleading caption candidates that have low distinctiveness scores can be effectively filtered out.


\paragraph{Uniqueness.} First, we demonstrate how to measure a description \textbf{uniquely} referring to its target.
We calculate the CLIP score between a caption and all masks and then divide the CLIP score for a target mask by the highest CLIP score among all other masks.
This is formulated as:
\begin{align}
    \text{UoS}_{i} = \frac{ \theta(x_{i}, \mathbf{y})}    {\max_{j \in C \backslash \{x_i\}} \theta(x_{j}, \mathbf{y})},
\end{align}
where $\theta(x, \mathbf{y})$ is a function calculating the CLIP score between inputs.
$x_i$ and $\mathbf{y}$ are a cropped patch and a description for a target $i$.
$x_j$ is a cropped patch for other instance $j$, where $j \in C \backslash \{x_i\}$.
By dividing the CLIP score for the target mask with a given description by the maximum CLIP score among other masks, the UoS score reflects  the relevance of the caption to the target compared to other masks.
If this ratio is close to or lower than 1, it indicates that the caption may refer to at least one other unintended mask.
However, the limitation of uniqueness is its inadequate ability to capture captions that correctly point to the target, caused by the simple calculation of the CLIP score between visual and textual embeddings without any in-depth consideration of whether the correct target object is being described.
This issue, depicted in Fig.~\ref{fig:distinctiveness}(a), highlights the need for another term to capture a description that correctly points to a target.

\paragraph{Correctness.}
To assess whether a caption describes the correct target or non-targeted object, we again employ the CLIP score.
But this time, we focus on the masked region of objects, excluding the surrounding context, and on a target noun phrase that identifies the subject described in the caption.
Formally, the correctness score is computed as follows:
\begin{align}
    \text{CoS}_{i} = \theta(x_{i} \odot m_i, \mathcal{T}_{\text{NP}}(\mathbf{y})),
    \label{eq:cos}
\end{align}
where $\odot$ is a Hadamard product operation, $m_i$ is a segmentation mask for a target $i$, and $\mathcal{T}_{\text{NP}}(\cdot)$ is a function that extracts a target noun phrase, as in~\cite{global_local_clip}.


\paragraph{Distinctiveness.}
We finally define distinctiveness scores by extending uniqueness scores using the correctness term in Eq~\eqref{eq:cos} as follows:
\begin{align}
    \text{DoS}_i = \frac{ \text{CoS}_{i} \times \theta(x_{i}, \mathbf{y})}    {\max_{j \in C \backslash \{x_i\}} \text{CoS}_{j} \times \theta(x_{j}, \mathbf{y})}.
\end{align}
In this manner, our distinctiveness score effectively measures how well a description \textbf{uniquely} and \textbf{correctly} refers to the target mask relative to other masks.
We present examples of our distinctiveness score in Fig.~\ref{fig:distinctiveness}.
On each mask, we only keep the candidates whose distinctiveness score is above a threshold $\tau$ and these are paired with their corresponding mask to form our pseudo supervision.

\section{Implementation Details}
\label{sec:implementation}

We utilize SAM~\cite{sam} as a mask generator and CoCa~\cite{coca} trained only on web-crawled datasets (LAION-2B~\cite{laion} and CC3M~\cite{cc3m}) as a captioning model.
For distinctiveness-based text filtering, we adopt CLIP~\cite{clip} with ResNet-50~\cite{resnet} and set a threshold $\tau$ to 1.3.
We generate our pseudo supervision from 16,992 unlabeled images in the training set of RefCOCO+.
We evaluate our pseudo supervisions using two RIS models: CRIS~\cite{cris} and ETRIS~\cite{etris}.
In all our experiments, we follow their configurations without modifications.


\paragraph{Mask Generation.}
We extract segmentation masks using SAM~\cite{sam}.
It generates an excessive number of masks and results in 2,026,942 masks from 16,992 raw images, making a significant computational challenge.
To address this, we reduce the mask count from overwhelming $\sim$120 SAM masks per image to a manageable range of $\sim$5 masks by taking overlaid SAM masks with CutLER~\cite{cutler} masks, which is an unsupervised segmentation model.
The visualization of this process and the performance using CutLER masks are reported in the supplementary.



\paragraph{Text Generation.}
For each segmentation mask, we generate 4 distinct types of cropped patches, each with a different margin setting which is 0\%, 10\%, 20\%, and a special 0\% with the masked region of objects, influencing the contents of the generated caption.
For each cropped patch, we generate 11 caption candidates using a captioning model by employing three decoding strategies (Beam search and our distinctive caption sampling with fixed top-$k$ or top-$p$ words), each with different hyperparameters (top-$k$ words by $k\in \{5, 7, 9, 11, 13\}$ or top-$p$ words~\cite{top-p} by $p\in \{0.4, 0.5, 0.6, 0.7, 0.8\}$).
Therefore, we generated a total of 44 caption candidates, and those captions with distinctiveness scores below the threshold $\tau$ are discarded.
The temperature is used as a default value in CoCa. 

\section{Experiments}
\label{sec:experiments}
\subsection{Datasets and Metrics.}
In our experiments, we use RefCOCO~\cite{refcoco}, RefCOCO+~\cite{refcoco}, RefCOCOg~\cite{refcocog, refcocog_google} and PhraseCut~\cite{phrasecut} datasets.
RefCOCO, RefCOCO+, RefCOCOg, and PhraseCut provide 19,994, 19,992, 26,711, and 77,262 images and 50,000, 49,856, 54,822, and 354,496 annotated instances with multiple referring descriptions, respectively.
In particular, PhraseCut offers a large number of instance categories (3101 classes), in contrast RefCOCO datasets which handle only 80 categories.
The referring descriptions in each dataset have different characteristics.
RefCOCO provides positional cues of objects within the image such as \textit{left} and \textit{right}, it is banned in RefCOCO+, and RefCOCOg has longer and more complex sentences than others.
PhraseCut presents a structured text form that includes attributes, categories, and relationships.
For evaluation, standard evaluation metrics in RIS are used: oIoU (overall Intersection over Union) and mIoU (mean Intersection over Union).
oIoU tends to penalize more in cases of failure in larger instances, because it calculates the total intersection divided by the total union across all predictions and ground truths masks.
In contrast, mIoU treats each instance equally, regardless of its size as it computes the IoU for each individual sample and then averages these values across all samples. 

\renewcommand{\arraystretch}{1.2}
\begin{table*}[t]
    \scriptsize
    \centering
    \caption{oIoU comparision with zero-shot and weakly-supervised RIS methods. Our pseudo supervised RIS models perform a zero-shot evaluation on RIS benchmark datasets. U and G denote UMD and Google splits of RefCOCOg, respectively. S and T are segmentation and text supervision by humans. $\dagger$ denotes the method utilizing multiple foundation models (\eg, Stable diffusion, SAM, and CLIP) as ours.}
    \scalebox{0.93}{
    \begin{tabular}{llll|ccc|ccc|ccc}
    \hline
  \multirow{2}{*}{\begin{tabular}[c]{@{}l@{}}Sup.\\ Types\end{tabular}} & \multirow{2}{*}{\begin{tabular}[c]{@{}l@{}}Human\\ Label\end{tabular}} & \multirow{2}{*}{Methods} & \multirow{2}{*}{\begin{tabular}[c]{@{}l@{}}Mask\\ Gen.\end{tabular}}   & \multicolumn{3}{c|}{RefCOCO}  & \multicolumn{3}{c|}{RefCOCO+} & \multicolumn{3}{c}{RefCOCOg} \\ \cline{5-13}
         & &  &  & val   & testA    & testB    & val   & testA    & testB    & val(U)  & test(U) & val(G) \\ \hline
 \multirow{2}{*}{\textit{Weakly}} & \multirow{2}{*}{T} &  SaG~\cite{shatter_and_gather} & & 30.40 & 29.75  &  31.24 &  25.27 & 24.46  & 26.04  & 27.12  &  - & - \\
     &&TRIS~\cite{tris} &   & 28.12  & 28.68  & 27.68  & 28.33  & 27.20 & 29.00 &  33.63 & 33.89  & 33.24 \\
    \hline
    \multirow{2}{*}{\textit{Zero-shot}} & \multirow{2}{*}{-} &   Global-Local~\cite{global_local_clip}  & SAM  & 24.55  & 26.00  & 21.03  & 26.62  & 29.99 & 22.23  & 28.92  &  30.48 & 30.19 \\
    & &   Ref-Diff$^\dagger$~\cite{ref_diff}   & SAM & 35.16  &  37.44 & \textbf{34.50}  & 35.56  & 38.66  & 31.40  & 38.62 & 37.50 & 37.82 \\    \cdashline{1-13}[1pt/1pt]
     \multirow{2}{*}{\textit{Pseudo}} &\multirow{2}{*}{-} &  Ours + CRIS   & SAM  & 35.45  & 39.69  & 30.37  & 37.55  & 41.52  & 31.71 & 37.90 & 39.64  & 39.32 \\
     & &   Ours + ETRIS   & SAM  &  \textbf{37.33} & \textbf{43.43}  & 31.90  & \textbf{40.19}  & \textbf{46.43}  & \textbf{33.63} & \textbf{41.63} & \textbf{43.52}  & \textbf{42.97} \\ \hline

     \multirow{2}{*}{\textit{Fully}} &\multirow{2}{*}{S+T} &  CRIS~\cite{cris} &    & 67.35 & 71.54 & 62.16 & 57.94 & 64.05 & 48.42 & 56.56 & 57.38 & -  \\ 
     & &  ETRIS~\cite{etris} &   & 69.36  & 72.68  & 63.68  & 58.59  & 66.93  & 50.14  & 58.62  & 60.12 & 55.56 \\
    \hline



    \end{tabular}
    \label{tab:main_results_oIoU}
    }
\end{table*}
\begin{table*}[t]
    \scriptsize
    \centering
    \caption{mIoU comparision with zero-shot and weakly-supervised RIS methods.}
    \scalebox{0.93}{
    \begin{tabular}{llll|ccc|ccc|ccc}
    \hline
  \multirow{2}{*}{\begin{tabular}[c]{@{}l@{}}Sup.\\ Types\end{tabular}} & \multirow{2}{*}{\begin{tabular}[c]{@{}l@{}}Human\\ Label\end{tabular}} & \multirow{2}{*}{Methods} & \multirow{2}{*}{\begin{tabular}[c]{@{}l@{}}Mask\\ Gen.\end{tabular}}   & \multicolumn{3}{c|}{RefCOCO}  & \multicolumn{3}{c|}{RefCOCO+} & \multicolumn{3}{c}{RefCOCOg} \\ \cline{5-13}
         & &  &  & val   & testA    & testB    & val   & testA    & testB    & val(U)  & test(U) & val(G) \\ \hline


     \multirow{3}{*}{\textit{Weakly}} &\multirow{3}{*}{T} &  Lee \textit{et al.}~\cite{weakly_ris_1} & & 31.06 & 32.30&30.11 &31.28 &32.11 & 30.13& 32.88 &- &- \\
    & & SaG~\cite{shatter_and_gather}  & & 34.76 &  34.58 & 35.01  & 28.48  & 28.60  & 27.98  & 28.87  & -  & - \\ 
    & & TRIS~\cite{tris} &   &  31.17 & 32.43  & 29.56  &  30.90 & 30.42 & 30.80 & 36.19  & 36.23  & 36.00 \\ \hline
    \multirow{2}{*}{\textit{Zero-shot}} & \multirow{2}{*}{-} &   Global-Local~\cite{global_local_clip} & SAM  & 31.83  & 32.93  & 28.64  & 34.97 & 37.11 & 30.61  & 40.66  & 40.94  & 41.68 \\
     &&   Ref-Diff$^\dagger$~\cite{ref_diff}   & SAM & 37.21  & 38.40  & \textbf{37.19}  & 37.29  & 40.51  & 33.01  & 44.02 & 44.51 & 44.26 \\    \cdashline{1-13}[1pt/1pt]
     \multirow{2}{*}{\textit{Pseudo}} &\multirow{2}{*}{-}&   Ours + CRIS   & SAM  & 39.75  & 44.77  &  32.99 & 42.15  & 46.28  & 34.54 & 43.73 & 43.37  & 43.78 \\
     & &  Ours + ETRIS   & SAM  & \textbf{41.05}  & \textbf{48.19} & 33.48 & \textbf{44.33}  & \textbf{51.42}  & \textbf{35.08} & \textbf{45.99} & \textbf{46.67}  & \textbf{46.80} \\ \hline
    \multirow{2}{*}{\textit{Fully}} &\multirow{2}{*}{S+T} &  CRIS~\cite{cris} &  & 70.34 & 73.41 & 65.72 & 62.13 & 67.68 & 53.42 & 60.30 & 60.29 & - \\ 
     & &  ETRIS~\cite{etris} &   & 71.06  & 74.11  & 66.66  & 62.23  & 68.51  & 52.79  & 60.28  & 60.42 & 57.86 \\

    \hline

    \end{tabular}
    \label{tab:main_results_mIoU}
    }
\end{table*}

\renewcommand{\arraystretch}{1}
\setlength{\tabcolsep}{6pt}
\begin{table}[t]
    \centering
    \footnotesize
    \caption{Cross-domain mIoU comparison between our pseudo-supervised and fully-supervised ETRIS models in two different domains, \eg RefCOCO and PhraseCut. Our pseudo-supervisions are generated using only train set images in the image dataset.}
    \scalebox{0.9}{
    \begin{tabular}{c|l|lll}
    \hline
         Supervision Types & Image Dataset & RefCOCO & RefCOCO+ & RefCOCOg \\  \hline
         \multirow{2}{*}{\textit{Fully}}& RefCOCO+& 64.54 & 62.23 & 60.35 \\ 
         & PhraseCut & 24.94 \tiny\textcolor{blue}{-39.60}  & 25.34 \tiny\textcolor{blue}{-36.89} & 26.41 \tiny\textcolor{blue}{-33.94} \\   
          \hline
         \multirow{2}{*}{ \makecell{ \textit{Pseudo} \\ (Ours) }}  & RefCOCO+& 41.05 & 44.33 & 45.99 \\
        &PhraseCut& 40.65 \tiny\textcolor{blue}{-0.40} & 43.25 \tiny\textcolor{blue}{-1.08} & 45.92 \tiny\textcolor{blue}{-0.07} \\

         \hline
    \end{tabular}
    }
    \label{tab:refcoco_out_domain}
\end{table}

\begin{table}[t]

    \centering
    \footnotesize
    \caption{Cross-domain and open-world results on PhraseCut.
    We generate our psuedo-supervisions using only train set images in the image dataset. \textit{Unseen} column denotes evaluation on a subset of object categories not included in RefCOCO datasets. $\dagger$ denotes the upper bound score of ETRIS where the training and testing domains are the same.} 
    \scalebox{0.9}{
    \begin{tabular}{c|p{2cm}|p{1.3cm}p{1.3cm}|p{1.3cm}p{1.3cm}}
    \hline
         \multirow{3}{*}{\makecell[l]{Supervision\\Types}} & \multirow{3}{*}{\makecell[l]{Image\\Dataset}} & \multicolumn{4}{c}{PhraseCut Test Set}  \\ \cline{3-6}
           & & \multicolumn{2}{c|}{mIoU} & \multicolumn{2}{c}{oIoU} \\ \cline{3-6}
           & & All & Unseen & All & Unseen \\ \hline
           \textit{Zero-shot~\cite{global_local_clip}} & N/A & 29.73 & 28.69\tiny\textcolor{blue}{-1.04} & 31.25 & 31.30\tiny\textcolor{red}{+0.05} \\ \hline
         
        \multirow{4}{*}{\textit{Fully}} & RefCOCO & 16.28 & 14.97\tiny\textcolor{blue}{-1.31} & 15.18 & 14.26\tiny\textcolor{blue}{-0.92} \\
          & RefCOCO+ & 16.14 & 14.94\tiny\textcolor{blue}{-1.20} & 14.93 & 14.11\tiny\textcolor{blue}{-0.72} \\
          & RefCOCOg & 13.03 & 11.92\tiny\textcolor{blue}{-1.11} & 11.80 & 10.96\tiny\textcolor{blue}{-0.84} \\ 
          & PhraseCut & \textcolor{gray}{38.10}$^\dagger$ & - & \textcolor{gray}{41.62}$^\dagger$ & - \\ \cdashline{1-6}[1pt/1pt]
          \multirow{2}{*}{\makecell{ \textit{Pseudo} \\ (Ours) }} & RefCOCO+ & 30.02 & 30.11\tiny\textcolor{red}{+0.09} & 28.68 & 29.14\tiny\textcolor{red}{+0.46} \\
           & PhraseCut & \textbf{33.70} & \textbf{33.95}\tiny\textcolor{red}{+0.25} & \textbf{32.75} & \textbf{33.52}\tiny\textcolor{red}{+0.77} \\ 
           
          
          \hline
    
    \end{tabular}
    }
    \label{tab:phrasecut}

\end{table}

\subsection{Results}
\paragraph{Main Results.}
In Table~\ref{tab:main_results_oIoU} and \ref{tab:main_results_mIoU}, we compare our pseudo-supervised method with zero-shot, weakly and fully supervised methods, in terms of oIoU and mIoU, respectively.
We perform a zero-shot evaluation on RefCOCO, RefCOCO+ and RefCOCOg using two RIS models, ETRIS~\cite{etris} and CRIS~\cite{cris}, trained with our pseudo supervisions generated from 16,994 raw images in the training set of RefCOCO+.
For a fair comparison with zero-shot methods, we re-implement Global-Local CLIP~\cite{global_local_clip} using SAM.
As shown in Table~\ref{tab:main_results_oIoU} and \ref{tab:main_results_mIoU}, our method significantly surpasses both zero-shot~\cite{global_local_clip, ref_diff}, and weakly supervised RIS methods~\cite{shatter_and_gather, tris, weakly_ris_1} by large margins.
We also report the scores of the fully supervised SoTA RIS models as the upper bound.

\begin{figure*}[t]
    \centering
    \begin{subfigure}{0.32\linewidth}
        \includegraphics[width=\linewidth]{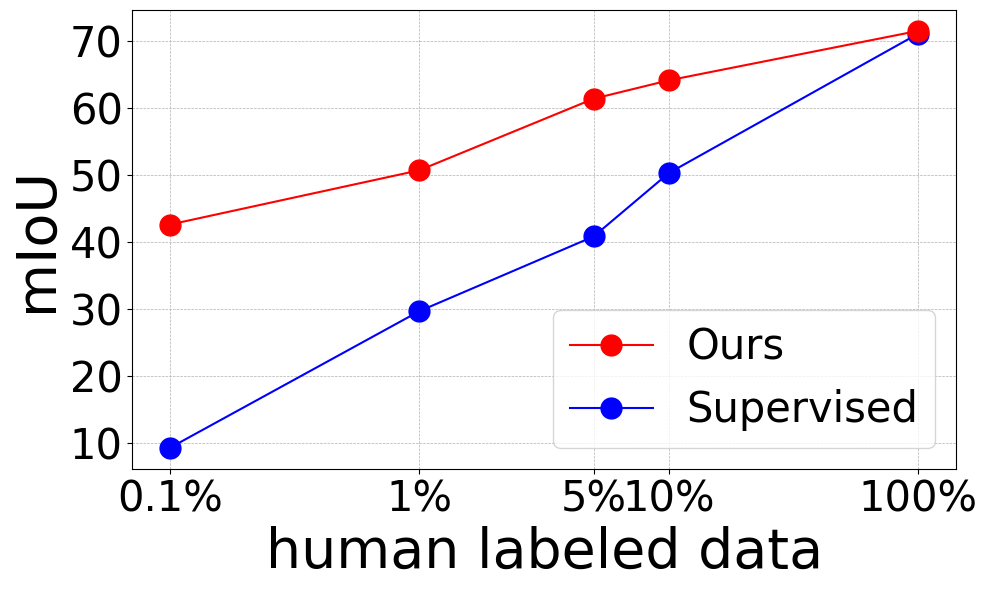}
        \caption{RefCOCO val}
        \label{fig:semi_refcoco}
    \end{subfigure}
    \begin{subfigure}{0.32\linewidth}
        \includegraphics[width=\linewidth]{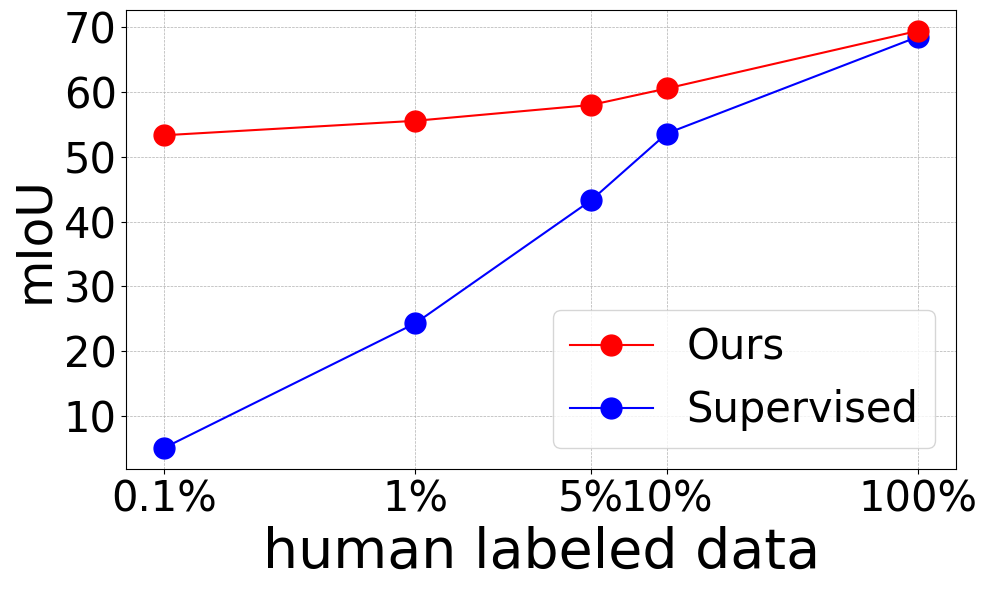}
        \caption{RefCOCO+ testA}
        \label{fig:semi_refcoco+}
    \end{subfigure}
    \begin{subfigure}{0.32\linewidth}
        \includegraphics[width=\linewidth]{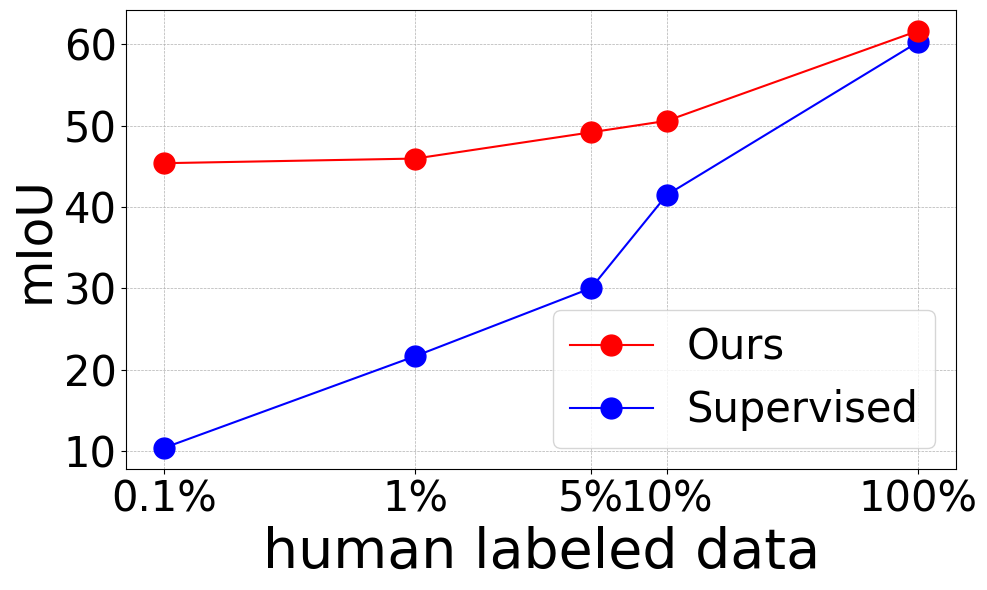}
        \caption{RefCOCOg test (U)}
        \label{fig:semi_refcocog}
    \end{subfigure}
    \caption{{mIoU results of two models: our Pseudo-RIS in a semi-supervised setting and a fully supervised model, across different volumes of human-labeled data. Our method uses both our pseudo-supervisions with varying proportions of human-labeled data, while the fully supervised method only relies on human labeled data. Both models are based on the ETRIS~\cite{etris}.} We also provide performance comparisons with other semi-supervised RIS methods~\cite{semi_ris_1, semi_ris_2, semi_safari} in our supplementary.}
    \label{fig:semi}
\end{figure*}

\paragraph{Cross-domain Results.}
The robustness of a model to work with consistent performance across domains is crucial for real-world applications.
In this context, we report the performances of both our pseudo-supervised and fully supervised ETRIS models trained on two different domains (\eg RefCOCO variants and PhraseCut) in Table~\ref{tab:refcoco_out_domain} and \ref{tab:phrasecut} for RefCOCO and PhraseCut datasets, respectively.
Our pseudo-supervisions are generated from raw images in a train set of the image dataset.
We observe that the performances of the supervised model trained with human-labeled data from PhraseCut or RefCOCO datasets drop significantly when evaluated on different domains.
This performance degradation is due to domain gaps~\cite{domain_ris} caused by differences in textual styles and object categories handled in each dataset. 
In contrast, our pseudo-supervised model shows consistent performances across different domains, validating a level of robustness superior to that of the fully supervised model trained with manual annotations.

\paragraph{Open-world Results.}
To validate the generalization ability of our Pseudo-RIS model compared to the fully supervised one, we extend our zero-shot evaluation to an unseen domain, PhraseCut~\cite{phrasecut}, as shown in Table~\ref{tab:phrasecut}.
Since PhraseCut has a larger number of instance categories (\ie 3,103 classes) than RefCOCO datasets (80 classes from MS-COCO~\cite{coco}), we choose PhraseCut to validate the generalization ability in open-world setting.
In Table~\ref{tab:phrasecut}, \textit{Unseen} column denotes when evaluated on a subset of instance categories not included in RefCOCO datasets as in \cite{global_local_clip}.
Note that the fully supervised model shows a significant drop in performance on the unseen subset, whereas our pseudo-supervised models are robust and even improve performance on this subset.
These results highlight the superior generalization ability of our pseudo-supervised model in light of the open-world nature of RIS.

\paragraph{Results on Semi-supervised Learning.}

Fig.~\ref{fig:semi} illustrates a comparison between our method and a supervised approach in a semi-supervised setting, both using the ETRIS model.
In this comparison, our method is first trained with our pseudo supervisions and then fine-tuned with labeled data, while the supervised method is trained solely on proportions of human-labeled data (0.1\%, 1\%, 5\%, 10\%, and up to 100\%).
Remarkably, our approach achieves promising results even when just 0.1\% labeled data.
As the proportion of labeled data increases to 1\%, 5\% and 10\%, we observe a consistent improvement in our performance, demonstrating the effectiveness of our method even with minimal labeled data.
These results highlight the potential of our method, especially in real-world scenarios where obtaining large amounts of labeled data is expensive and challenging.
Additionally, we provide a performance comparison with other semi-supervised RIS methods~\cite{semi_ris_1, semi_ris_2, semi_safari} in our supplementary materials.

\subsection{Ablation Study}
We conduct ablation studies on the val set of each RefCOCO dataset to validate the effectiveness of our method.
We adopt ETRIS as our base model trained with our pseudo supervisions in all experiments.


\paragraph{Impact of Proposed Components.}
In this ablation study, we analyze the impact of each component in our method, as illustrated in Table~\ref{tab:ablation_components}.
The results show that the integration of our \textit{distinctive caption sampling} into the na\"ive method significantly improves the performances across all datasets in terms of mIoU scores.
We also demonstrate that our \textit{distinctiveness-based text filtering} improves mIoU scores.
Furthermore, combining \textit{distinctiveness-based text filtering} with  \textit{distinctive caption sampling} leads to further improvements in mIoU scores.
These results indicate that our two strategies are more effective than the na\"ive method in generating distinctive captions for each mask.


    

\begin{table}[t]
    \centering
    \small
    \caption{mIoU results with different components of our approach. DCS and filtering denote our distinctive caption sampling and distictiveness-based text filtering methods, respectively.}
    \scalebox{0.9}{
    \begin{tabular}{cc|lll}
    \hline
    DCS & Filtering & RefCOCO & RefCOCO+ & RefCOCOg \\ \hline
        &     & 37.96  & 39.87 & 42.08 \\
    \checkmark & & 40.01 \scriptsize \textcolor{red}{+2.05} & 43.27 \scriptsize \textcolor{red}{+3.40} & 44.46 \scriptsize \textcolor{red}{+2.38} \\
    & \checkmark & 40.40 \scriptsize \textcolor{red}{+2.44} & 43.40 \scriptsize \textcolor{red}{+3.53} & 45.09 \scriptsize \textcolor{red}{+3.01} \\
    \checkmark & \checkmark & \textbf{41.05} \scriptsize\textcolor{red}{+3.09} & \textbf{44.33} \scriptsize\textcolor{red}{+4.46} & \textbf{45.99} \scriptsize \textcolor{red}{+3.91} \\ \hline
    
    \end{tabular}
    }
    \label{tab:ablation_components}
\end{table}

\begin{table}[t]
    \centering
    \footnotesize
    \caption{mIoU results of ETRIS model with full supervisions, GT segmentation masks with our generated pseudo text supervisions, and solely our pseudo-supervisions. S and T indicate GT segmentation masks and GT texts used for training. }
    \scalebox{0.9}{    
    \begin{tabular}{l|c|ccc|ccc}
    \hline
     \multirow{2}{*}{\makecell[l]{Supervision\\Types}} & \multirow{2}{*}{\makecell[c]{Human\\Label}}  & \multicolumn{3}{c|}{RefCOCO+} & \multicolumn{3}{c}{RefCOCOg} \\ \cline{3-8}
     & & val & testA  & testB & val(U) & test(U) & val(G) \\ \hline

     \multirow{1}{*}{\textit{Fully}} & S+T & 62.23 & 68.51 & 52.79 & 60.28 & 60.42 & 57.86 \\ \cdashline{1-8}[1pt/1pt]
     \multirow{2}{*}{\makecell[l]{\textit{Pseudo}\\(Ours)}} & S & 52.54 & 60.24 & 42.43 & 54.66 & 54.78 & 55.82 \\
      & - & 44.33 & 51.42 & 35.08 & 45.99 & 46.67 & 46.80 \\
    \hline
    
    \end{tabular}
    }
    \label{tab:mask_generator}
\end{table}

\paragraph{Effects on GT segmentation masks.}
In Table~\ref{tab:mask_generator}, we show the impact of quality on segmentation masks using ETRIS as a base model. 
By using GT masks to generate our distinctive text supervisions, we achieve remarkable performances that are close to a fully supervised model.
These results highlight the powerful ability of our method to generate distinctive expressions for RIS.

\paragraph{Ablation on threshold $\tau$.}

\begin{figure}[t]
    \centering
    \begin{subfigure}{1\linewidth}
        \centering
        \includegraphics[width=1\linewidth]{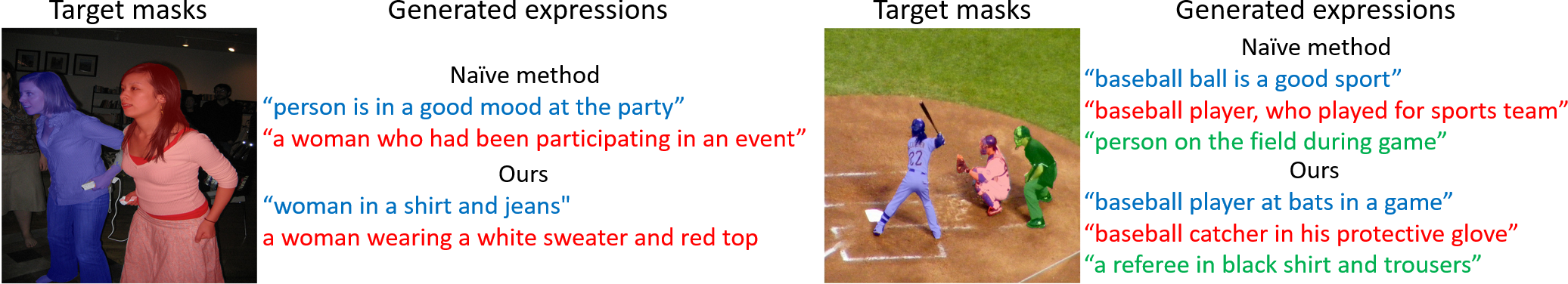}
        \caption{Comparison of our generated expressions with the na\"ive method.}
        \label{fig:qualitative}
    \end{subfigure}
    \hfill
    \begin{subfigure}{1\linewidth}
        \centering
        \includegraphics[width=1\linewidth]{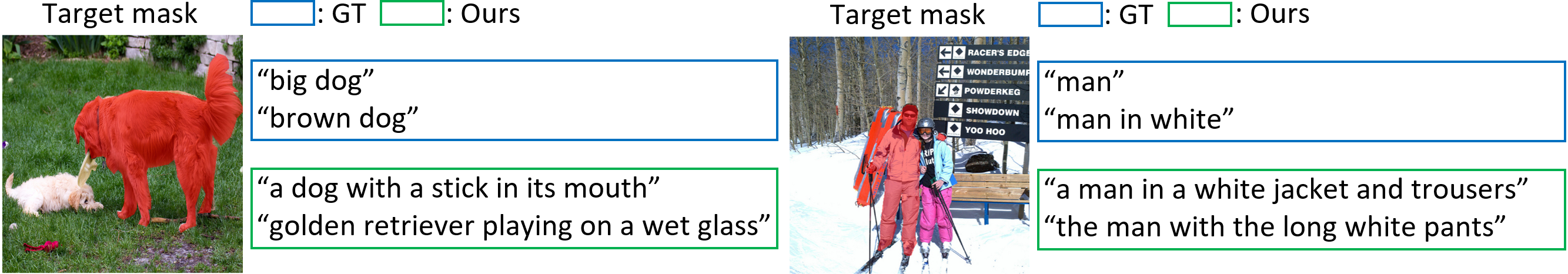}
        \caption{Comparison of our generated expressions with GT expressions.}
        \label{fig:qualitative_with_gt}
    \end{subfigure}
    \caption{Qualitative analysis of our generated expressions compared to the na\"ive methods and GT captions.}
    \label{fig:qual_anylsis}
    \label{fig:qualitative_naylsis}
\end{figure}

We conduct an ablation study on the threshold $\tau$, which is crucial for filtering out misleading captions through our \textit{distinctiveness metric}.
The mIoU scores under various threshold $\tau$ values are reported in our supplementary material.
We choose $\tau = 1.3$ for all datasets.
\paragraph{Qualitative Analysis.}
We visualize the expressions generated by our method and those generated by the na\"ive method in Fig.~\ref{fig:qualitative}, as well as compare ours with GT expressions in RefCOCO datasets in Fig.~\ref{fig:qualitative_with_gt}.
A na\"ive method generates non-distinctive captions, which make it hard to distinguish the target object well from other objects.
In contrast, our method produces distinctive captions describing the details of the target mask.
Compared to GT expressions, ours provide more variety of visual concepts in the target mask, making the RIS models robust and general to different domains.

\begin{figure}[t]
    \centering
    \begin{subfigure}{1\linewidth}
        \centering
        \includegraphics[width=0.9\linewidth]{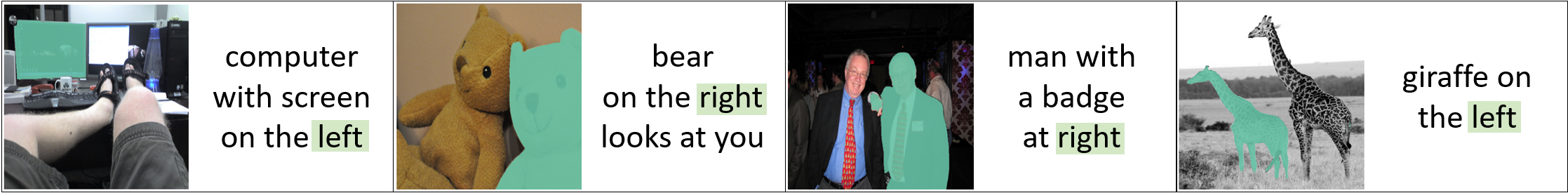}
        \caption{Examples of our pseudo-captions with positional words.}
        \label{fig:qualitative_with_position}
    \end{subfigure}
    \hfill
    \begin{subfigure}{1\linewidth}
        \centering
        \includegraphics[width=0.9\linewidth]{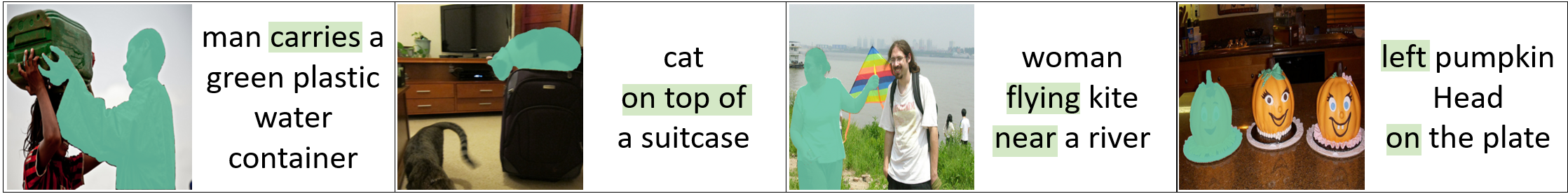}
        \caption{Examples of our captions with relations between objects.}
        \label{fig:qualitative_with_relation}
    \end{subfigure}
    \caption{Qualitative analysis of our generated expressions with positions and relations.}

    \label{fig:enter-label}
\end{figure}






\paragraph{Limitations and Discussions.}
The proposed method is not explicitly designed to generate phrases describing object positions (\eg \textit{left}, \textit{right}) or relations between objects (\eg \textit{on top of}).
However, we find that it can still produce such phrases to some extent as showcased in Fig~\ref{fig:qualitative_with_position} and \ref{fig:qualitative_with_relation}.
This capability is due to two factors: (1) we crop patches with margins that include contextual information around the target, and (2) we employ a captioning model trained on web datasets that include positional and relational words.
We plan to explore this matter further in future work.


\label{sec:limitation}

\paragraph{Social Impacts.}
Since our method relies on foundation models that are trained on large-scale datasets, it also inherits their limitations, such as dataset bias and unfairness issues. 
Specifically, we utilize the image captioning model trained with non-curated web-crawled large-scale datasets, and it may generate captions that include discriminatory or biased languages present in the dataset.
This highlights the need to carefully monitor the generated expressions for identifying and correcting biased or harmful content.





\section{Conclusion}
\label{sec:conclusion}
In this paper, we propose a novel framework, Pseudo-RIS, to generate pseudo-supervisions for RIS, where two techniques are introduced: (1) generating multiple candidates with distinctive words, and (2) filtering out the misleading captions, resulting in distinctive captions, like manual supervisions.
Our approach outperforms SoTA weakly and zero-shot approaches, even fully supervised methods on an unseen domain.
Furthermore, we show effectiveness of our method in a semi-supervised setting with a few labeled data.





\paragraph{Acknowledgements.}
This work was supported by the IITP grants (No.2019-0-01842~(3\%), No.2021-0-02068~(2\%), No.2022-0-00926~(50\%), No.2024-2020-0-01819~(5\%)) funded by MSIT, Culture, Sports and Tourism R\&D Program of the KOCCA grant (RS-2024-00345025~(5\%)) funded by MCST, and the GIST-MIT Research Collaboration grant~(35\%) funded by GIST, Korea.

%
%
\bibliographystyle{splncs04}
\bibliography{egbib}

\title{Pseudo-RIS: Distinctive Pseudo-supervision Generation for Referring Image Segmentation \\ - \textit{Supplementary Materials} -} 


\titlerunning{Pseudo-RIS}

\author{Seonghoon Yu\inst{1} 
\quad 
Paul Hongsuck Seo\inst{2}$^\dagger$ 
\quad 
Jeany Son\inst{1}$^\dagger$
}

\authorrunning{Yu et al.}

\institute{\hspace{-0.15cm}AI Graduate School, GIST, Korea \and
\hspace{-0.1cm}Department of Computer Science and Engineering, Korea University, Korea
\\
\email{seonghoon@gm.gist.ac.kr \quad phseo@korea.ac.kr \quad jeany@gist.ac.kr}
\\
\url{https://github.com/Seonghoon-Yu/Pseudo-RIS} 
}


\maketitle

\appendix
\label{sec:appendix}

\appendix
\label{sec:appendix}

\setcounter{table}{6}
\setcounter{figure}{6}

\renewcommand{\thefootnote}{}
\footnotetext{$^\dagger$Corresponding authors.}
\renewcommand{\thefootnote}{\arabic{footnote}}

\section{Pseudo-supervision Generation on Additional Datasets}
A key advantage of our proposed Pseudo-RIS is its ability to create pseudo RIS annotations from unlabeled images in various datasets.
Table~\ref{sup_tab:different datasets} shows mIoU results with our pseudo-supervisions generated from raw images of different datasets (PascalVOC~\cite{pascalvoc}, Flickr30k~\cite{flickr30k}, PhraseCut, and RefCOCO+), as well as their combination.
Additionally, we report the mIoU results of ETRIS trained separately on individual RefCOCO datasets in Table~\ref{sup_tab:each_refcoco}.
From these results, we highlight the effectiveness of our method in expanding the scale of our pseudo annotations and the generalization ability of our pseudo-supervisions.



\begin{table}[h]
    \centering
    \footnotesize
    \caption{mIoU results with different datasets and all combined.}
    \scalebox{0.96}{
    \begin{tabular}{@{\hspace{7pt}}l|@{\hspace{7pt}}c@{\hspace{7pt}}c@{\hspace{7pt}}c@{\hspace{7pt}}c@{\hspace{7pt}}}
    \hline
        \multirow{2}{*}{\makecell[l]{Image\\Dataset}} & RefCOCO & RefCOCO+ & RefCOCOg & PhraseCut \\
        & val & val & val(U) & test \\ \hline
        PascalVOC & 37.41 & 39.08 & 41.85 & 31.30 \\
        Flickr30k & 38.59 & 41.49 & 42.48 & 32.39 \\
        PhraseCut & 40.65 & 43.25 & 45.92 & 33.70 \\ 
        RefCOCO+ & 41.05 & 44.33 & 45.99 & 30.02 \\
        \cdashline{1-5}[1pt/1pt]
        All combined & \textbf{42.18} & \textbf{45.75} & \textbf{47.40} & \textbf{34.02} \\
    \hline
    \end{tabular}
    }
    \label{sup_tab:different datasets}
\end{table}

\begin{table}[h]
    \centering
    \footnotesize
    \caption{mIoU results on individual RefCOCO variants.}
    \scalebox{0.96}{
    \begin{tabular}{@{\hspace{7pt}}l@{\hspace{7pt}}|c@{\hspace{7pt}}c@{\hspace{7pt}}c@{\hspace{7pt}}}
    \hline
        \multirow{2}{*}{\makecell[l]{Image\\Dataset}} & RefCOCO & RefCOCO+ & RefCOCOg \\
        & val & val & val(U) \\ \hline
        RefCOCO & 40.20 & 43.37 & 44.78 \\
        RefCOCO+ & 41.05 & \textbf{44.33} & 45.99 \\
        RefCOCOg & \textbf{41.34} & 44.06 & \textbf{46.98} \\ 
    \hline
    \end{tabular}
    }
    \label{sup_tab:each_refcoco}
\end{table}

\section{Impact of Mask Quality}
To show the impact of varying mask quality on the Pseudo-RIS framework, we evaluate the performance of our pseudo-supervised ETRIS~\cite{etris} model and the zero-shot RIS method (Global-Local CLIP~\cite{global_local_clip}) using different segmentation models (CutLER~\cite{cutler}, SAM~\cite{sam}, and GT masks) for mask extraction, as shown in Table~\ref{sup_tab:different_mask_generators}.
The zero-shot RIS method applies the extracted masks during inference, while our approach uses them during training.
Despite having a disadvantage when using GT masks, as the zero-shot RIS method can select the best mask from all GT object masks in an image, our method still outperforms the zero-shot RIS method with GT masks.
Despite this disadvantage, our method surpasses the zero-shot RIS method when using GT masks.
Remarkably, when utilizing the unsupervised segmentation model, CutLER~\cite{cutler}, our method achieves superior performance compared to the zero-shot RIS method that employs SAM.
\begin{table}[h]
    \centering
    \footnotesize
    \caption{mIoU results using various segmentation models for mask extraction compared to the zero-shot RIS method. Notice that, the zero-shot RIS method uses the extracted masks during the inference, whereas ours uses these masks in the training.}
    \begin{tabular}{@{\hspace{7pt}}l@{\hspace{7pt}}c@{\hspace{7pt}}l@{\hspace{7pt}}|c@{\hspace{7pt}}c@{\hspace{7pt}}c@{\hspace{7pt}}}
    \hline
    Mask & Inference & Methods & RefCOCO & RefCOCO+ & RefCOCOg \\ \hline
    \multirow{2}{*}{\textit{CutLER}}& \ding{51} & \textit{Zero-shot}~\cite{global_local_clip}  & 27.35 & 28.34 & 33.23 \\
    & \ding{55} & \textit{Pseudo} \scriptsize (Ours) & \textbf{37.40} & \textbf{39.42} & \textbf{42.63} \\ \hline
    \multirow{2}{*}{\textit{SAM}}& \ding{51} & \textit{Zero-shot}~\cite{global_local_clip}  & 31.83 & 34.97 & 40.66 \\
    & \ding{55} & \textit{Pseudo} \scriptsize (Ours)& \textbf{41.05} & \textbf{44.33} & \textbf{45.99} \\ \hline    
    \multirow{2}{*}{\textit{GT}}& \ding{51}  & \textit{Zero-shot}~\cite{global_local_clip}  & 37.05 & 42.59 & 51.01 \\
    & \ding{55} & \textit{Pseudo} \scriptsize (Ours) & \textbf{47.22} & \textbf{52.54} & \textbf{54.66} \\ \hline
    
    \end{tabular}
    \label{sup_tab:different_mask_generators}
\end{table}

\section{Analysis on Word Coverage}
During the RIS model training, exposing it to a broad and diverse vocabulary is crucial for effective performance in various domains or open-world scenarios.
We analyze the scope of the word vocabulary covered in our generated expressions compared to ground-truth expressions on RefCOCO datasets in Table~\ref{sup_tab:word_vocap}.
Our expressions cover an extensive vocabulary of 37,588 words, surpassing that of RefCOCO datasets.
This breadth and diversity of vocabulary in ours contribute to our superior performance over the manual labels in cross-domain and open-world settings, as shown in Sec 5.2 of the main manuscript.
\begin{table}[h]
    \centering
    \footnotesize
    \caption{The scope of word vocabulary in our expressions and GT expressions.}
    \begin{tabular}{@{\hspace{7pt}}c@{\hspace{7pt}}|c@{\hspace{7pt}}c@{\hspace{7pt}}c@{\hspace{7pt}}c@{\hspace{7pt}}}
    \hline
         & Ours &RefCOCO & RefCOCO+ & RefCOCOg \\ \hline
       \# of Vocabulary & \textbf{37,588} & 9,342 & 11,090 & 11,710 \\
    \hline
    \end{tabular}
    \label{sup_tab:word_vocap}
\end{table}

\section{Advantages over simply using MLLM}
\begin{figure}[h]
    \begin{subfigure}[b]{1\textwidth}
    \centering
    \includegraphics[width=0.94\textwidth]{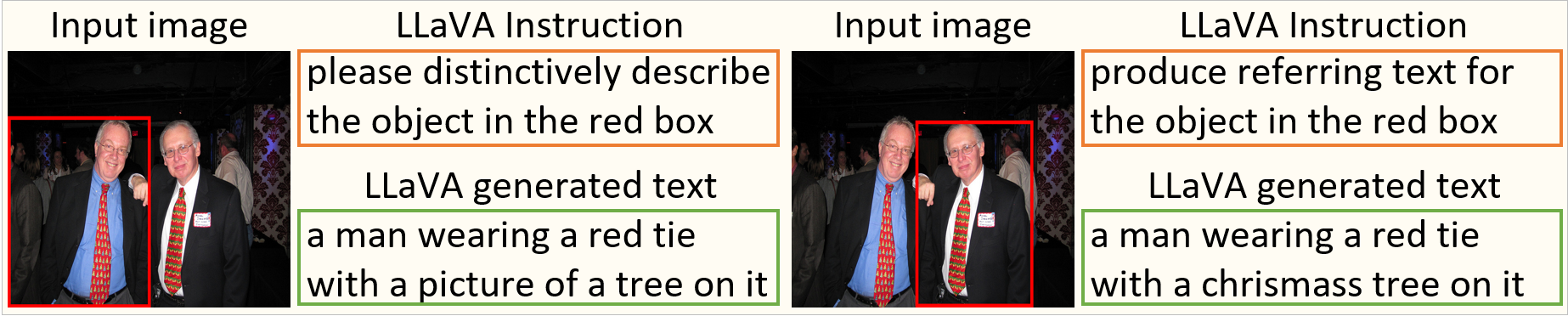}
    \caption{Examples of the input image with marked and an instruction for LLaVA, and its output.}        
    \end{subfigure}
    \begin{subfigure}[b]{1\textwidth}
    \centering
    \includegraphics[width=0.94\textwidth]{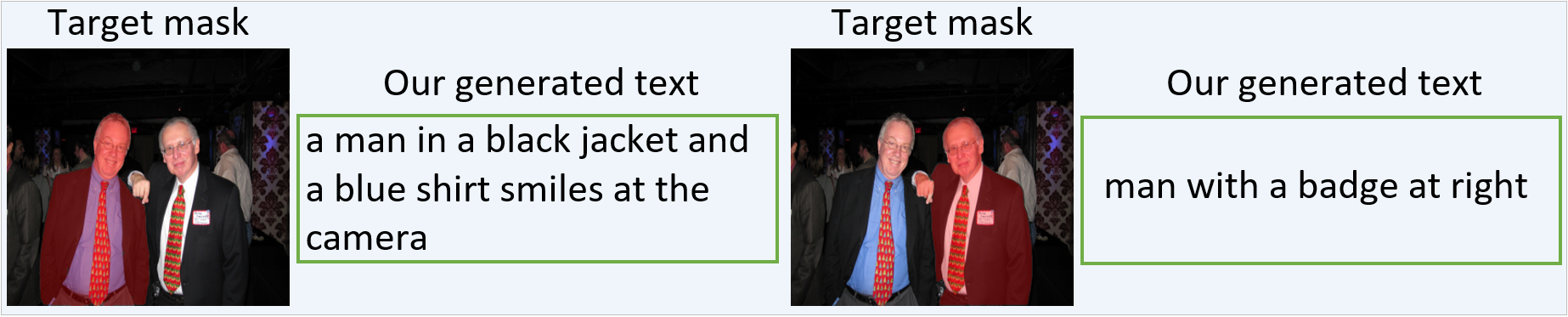}
    \caption{Examples of our generated referring expressions.}        
    \end{subfigure}
    \caption{Qualitative analysis of our generated expressions compared to that of simply using LLaVA, where we follow InstructDET. We observe that LLaVA produces massive hallucinations, as mentions in InstructDet (they fine-tune LLaVA on RefCOCO to handle it). Our generated captions are more distinctive than that of LLaVA.}
    \label{fig:llava}
\end{figure}

To show the impact of our work over MLLMs~\cite{mllm_7,mllm_8} in generating high-quality referring text, we compare our generated text with that of LLaVA~\cite{llava} in terms of performance and qualitative examples in Table~\ref{supple_tab:llava} and Fig~\ref{fig:llava}, respectively.
For this analysis, we follow the concurrent study, InstructDET~\cite{recent_pseudo_6}, to generate unique referring text using LLaVA.
Since InstructDET is designed for a REC task, we use boxes that cover SAM's masks.
The performance of simply using MLLMs is inferior to our methods.
The issues of using LLaVA na\"ively were already highlighted in the InstructDet paper: (1) it inaccurately describes specific objects and (2) produces massive hallucinations.
As a result, they fine-tuned LLaVA on REC datasets, implying challenges of using MLLMs to generate high-quality referring data without modifications.
Moreover, our method has several advantages over pixel-level MLLMs~\cite{mllm_1,mllm_2,mllm_3,mllm_4,mllm_5,mllm_6,mllm_9, mllm_10, mllm_11, mllm_12, mllm_13, mllm_14, mllm_15, mllm_16, mllm_17}: (1) it is a training-free method, adjusting the output logits (word probabilities) from a frozen captioning model, (2) it introduces a newly designed metric for the distinctiveness of captions, and (3) it is memory-efficient.


\begin{table}[h]
    \centering
    \footnotesize
    \caption{mIoU results of simply using LLaVA following InstructDET. We evaluate generated pseudo-supervisions using ETRIS.}
    \begin{tabular}{@{\hspace{7pt}}l@{\hspace{7pt}}c@{\hspace{7pt}}|c@{\hspace{7pt}}c@{\hspace{7pt}}c@{\hspace{7pt}}}
    \hline
    Models & Params  & RefCOCO & RefCOCO+ & RefCOCOg \\ \hline
    LLaVA  & 13B & 20.51 & 20.84 & 20.42 \\
    Ours (CoCa) & 2.1B & 41.05 & 44.33 & 45.99 \\ \hline
    \end{tabular}
    \label{supple_tab:llava}
\end{table}


\section{Standard Decoding Strategy}
In this section, we briefly demonstrate the standard decoding strategies that are commonly used in text generation tasks, as mentioned in Sec 3.2 of the main manuscript.
Basically, they select the next word $y_t$ from the probabilities of the words $P(y|y_{< t})$, based on the previously generated words $y_{< t} = \langle y_1, \dots, y_{t-1} \rangle$.
The specific process for selecting the next word is different for each decoding method and how it is processed can significantly influence the quality and diversity of the generated text.





\begin{itemize}
    \item \textbf{Greedy Decoding:} At each step $t$, select the next word $y_t$ with the maximum probability, given the previously generated words. 
    Formally, $y_{t}$ $=$ $\arg \max_{y \subset V}$ $P(y|y_{< t})$.
    Where $V$ is the vocabulary.

    \item \textbf{Beam Search~\cite{beam_search}:}
    It simultaneously explores the multiple sequences and keeps only the top $k$ most promising ones, which are called \textit{Beams}.
    At each step $t$, it extends each sequence within the beams by adding the next word $y_t$ based on the probability of the next word for each sequence.
    Then, it keeps only the top $k$ sequences based on the cumulative probability of the words in each sequence.
    Finally, the sequence with the highest cumulative probability of all beams is selected as the final output.

    \item \textbf{Pure Sampling:} This is the most basic form of sampling-based decoding. At each step $t$, the next word $y_{t+1}$ is directly sampled from $P(y|y_{< t})$, where $y \in V$.
    
    \item \textbf{Top-k Sampling~\cite{top-k}:} Restrict the vocabulary $V$ to top $k$ words with the highest probabilities and it samples the next word from this restricted vocabulary $V_k$.
    Formally, $y_{t}$ $\sim$ $P(y_{t}|y_{< t})$, where $y \in V_k$.

    \item \textbf{Top-p Sampling~\cite{top-p}:} Instead of fixed $k$ words, a dynamic subset of vocabulary $V_p$ with top $p$ words is chosen such that the cumulative probability exceeds a predefined threshold $p$.
    In formal terms, $V_p$ is such that $\sum_{y \in V_p} P(y) \geq p$.
    The next token $y_{t+1}$ is sampled from this subset $V_p$.

\end{itemize}
Our proposed \textit{distinctive caption sampling}, which is a sampling-based decoding method, is designed to be compatible with standard decoding strategies.
We implement our decoding method in conjunction with top-$k$ and top-$p$ sampling by restricting the vocabulary to either the top-$k$ or top-$p$ words.

\paragraph{Limitations of Standard Decoding Methods.}
These standard decoding methods are mainly proposed to handle the NLP problems (\eg repetition, perplexity, diversity, and coherence) in text generation tasks.
Therefore, employing them directly in the image captioning model to generate distinctive captions for the RIS annotations results in non-distinctive captions that do not distinctively refer to the target mask, apparently when similar objects are present in the image.
This is because they rely solely on the provided target cropped patch $x_i$ to select the next word as $P(y_t|y_{\leq t}, x_i)$, without considering other objects within the image.
In contrast, the proposed \textit{Distinctive Caption Sampling} considers other objects in the same image to sample the next word focused on the target.

\section{Details of SAM}
\begin{figure}[h]
    \centering
    \begin{subfigure}[b]{0.31\textwidth}
    \centering
        \includegraphics[width=1\textwidth]{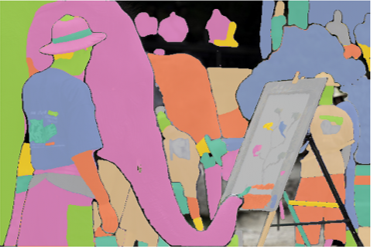}
        \caption{SAM masks ($\sim$ 120 masks)}
        \label{fig:sam}
    \end{subfigure}
    \hfill
    \begin{subfigure}[b]{0.31\textwidth}
        \centering
        \includegraphics[width=1\textwidth]{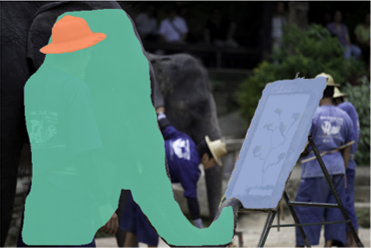}
        \caption{CutLER masks (3 masks)}
        \label{fig:cutler}
    \end{subfigure}
    \hfill
    \begin{subfigure}[b]{0.31\textwidth}
        \centering
        \includegraphics[width=1\textwidth]{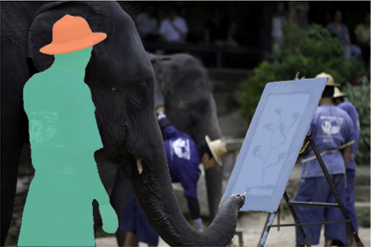}
        \caption{Overlaid SAM masks (3)}
        \label{fig:overlaid}
    \end{subfigure}
    
    \caption{Detailed processing of SAM masks.}
    \label{fig:detail_sam}
\end{figure}

\paragraph{Detailed Processing of SAM Masks.}
As mentioned in the implementation details (Section 4) of the main manuscript, we report the detailed processing of SAM masks in Fig.~\ref{fig:detail_sam}.
As shown in Fig.~\ref{fig:sam}, SAM extracts the overwhelming number of masks ($\sim$ 120 masks) per image, resulting in a substantial computational resource requirement to generate captions on each mask.
To deal this, we use overlaid SAM masks (Fig.~\ref{fig:overlaid}) with CutLER masks (Fig~\ref{fig:cutler}) to reduce the excessive number of SAM masks (Fig.~\ref{fig:sam}) to a manageable range of $\sim 5$ masks.
For details, we select SAM masks with the highest IoU with all CutLER masks ($\sim 5$ masks per image) from all SAM masks ($\sim120$ masks per image) to filter out small over-segmented SAM masks.
Note that, CutLER~\cite{cutler} is an unsupervised segmentation model and we report the performance results using CutLER masks in Table~\ref{sup_tab:different_mask_generators}.


\section{Analysis on the Distinctiveness Scores}
\paragraph{Comparision with Uniqueness and Correctness.}
To validate the effectiveness of the proposed metric, \textit{distinctiveness}, in the text filtering, we compare it with different metrics, \textit{uniqueness} and \textit{correctness} in Table~\ref{sup_tab:distinctiveness_compare}.
We observe that using the distinctiveness metric to filter out ambiguous and imprecise captions results in superior performance on the validation set of RefCOCO datasets.
These results demonstrate that the distinctiveness metric, which is designed to consider both factors (uniqueness and correctness), is more effective at eliminating misleading captions than considering each factor alone.





\begin{table}[h]
    \centering
    \footnotesize
    \caption{mIoU results with different metrics for the text filtering.}
    \begin{tabular}{@{\hspace{7pt}}l@{\hspace{7pt}}c|c@{\hspace{7pt}}c@{\hspace{7pt}}c@{\hspace{7pt}}}
    \hline
       Metric & Filtering & RefCOCO & RefCOCO+ & RefCOCOg \\ \hline
       -  & \ding{55} & 40.01 & 43.27 & 44.46 \\ \cdashline{1-5}[1pt/1pt]
      Uniqueness & \ding{51} & 40.55 & 43.39 & 45.49 \\
       Correctness  & \ding{51} & 40.65 & 43.67 & 45.88 \\
       Distinctiveness & \ding{51} & \textbf{41.05} & \textbf{44.33} & \textbf{45.99} \\
    \hline
    \end{tabular}
    \label{sup_tab:distinctiveness_compare}
\end{table}

\paragraph{Relation to Distinctive Caption Sampling.}
In Table~\ref{sup_tap:relation_to_distinctiveness}, we show that using our decoding strategy (distinctive caption sampling) to generate captions enhances the average distinctiveness scores of all generated captions.
This indicates that our decoding method is more effective in generating distinctive captions than the na\"ive decoding, which refers to standard decoding strategies (top-$k$ and top-$p$ sampling).
\begin{table}[h]
    \centering
    \small
    \caption{The average scores of distinctiveness for all captions generated by our decoding (distinctive caption sampling) compared to the na\"ive decoding (top-$k$ and top-$p$ sampling).}
    \begin{tabular}{@{\hspace{7pt}}l|c@{\hspace{0.5em}}c@{\hspace{0.5em}}c@{\hspace{7pt}}}
    \hline
         Metric & Na\"ive decoding & & Our decoding \\ \hline
         Distinctiveness & 1.161 & $\rightarrow$ & 1.260 \\
    \hline
    \end{tabular}
    \label{sup_tap:relation_to_distinctiveness}
\end{table}

\section{Ablation on distinctive caption sampling}

\paragraph{Weighted Sum to Aggregate Word Distributions.}
We conduct an ablation study on the proposed \textit{distinctive caption sampling} with another formation, where we use a weighted sum to aggregate the word distributions of other patches, instead of an average used in the original version.
This is formulated as follows:
\begin{align}
    &P(y_{t}|y_{< t}, x_i, C)\nonumber\\
    &= \sigma \big( P(y_{t}|y_{< t}, x_i) - \hspace{-0.4cm} \sum_{ x_j \in C \backslash \{x_i\}} \hspace{-0.2cm}  w_{ij}P(y_{t}|y_{< t}, x_j)  \big),\\
    &\text{where, } w_{ij} = \frac{s_{ij}}{\sum_{j \neq i} s_{ij}}, \nonumber
\end{align}
In Table~\ref{sup_tab:weighted_sum}, we report mIoU results using a weighted sum for aggregating the word distributions from other patches compared to an average (original version).
We observe that using an average as an aggregation outperforms using a weighted sum on all datasets.
\setlength{\dashlinedash}{1pt} 
\setlength{\dashlinegap}{1pt}  

\begin{table}[h]
    \centering
    \caption{mIoU results using different aggregations in our distinctive caption sampling to aggregate the word distributions from other patches to penalize that of the target.}
    \footnotesize
    \begin{tabular}{@{\hspace{7pt}}l@{\hspace{7pt}}|c@{\hspace{7pt}}c@{\hspace{7pt}}c@{\hspace{7pt}}}
    \hline
       Aggregation  &  RefCOCO & RefCOCO+ & RefCOCOg \\ \hline
       Weighted sum  & 39.66 & 42.63  & 43.56 \\
       Average (Ours)& \textbf{41.05} & \textbf{44.33} & \textbf{45.99} \\
   \hline
    \end{tabular}
    \label{sup_tab:weighted_sum}
\end{table}

\section{Ablation on threshold $\tau$ in Text Filtering}
\begin{figure}[h]
    \centering
    \begin{subfigure}{0.29\linewidth}
        \includegraphics[width=\linewidth]{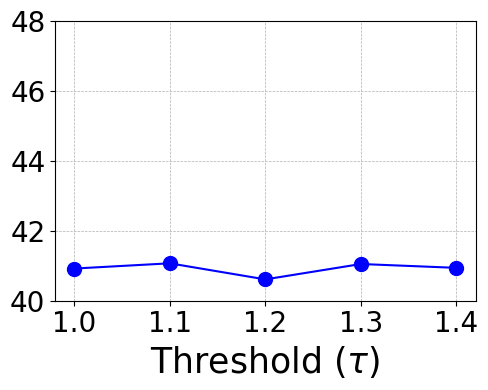}
        \caption{RefCOCO}
        \label{fig:threshold_refcoco}
    \end{subfigure}
    \hfill
    \begin{subfigure}{0.29\linewidth}
        \includegraphics[width=\linewidth]{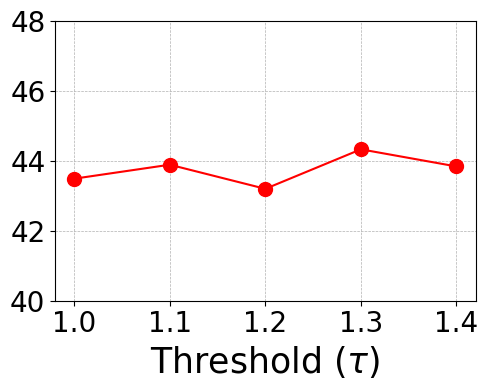}
        \caption{RefCOCO+}
        \label{fig:threshold_refcoco+}
    \end{subfigure}
    \hfill
    \begin{subfigure}{0.29\linewidth}
        \includegraphics[width=\linewidth]{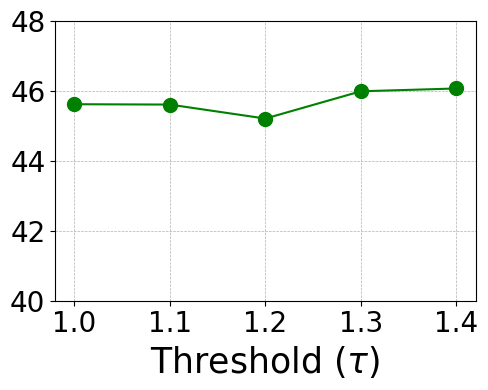}
        \caption{RefCOCOg}
        \label{fig:threshold_refcocog}
    \end{subfigure}
    \caption{mIoU results across different thresholds $\tau$ for the distinctiveness-based text filtering.}
    \label{fig:threshold}
\end{figure}

We report mIoU results under various threshold $\tau$ values to filter out non-distinctive captions, as mentioned in Sec 5.3 of the main manuscript.
Based on Fig.~\ref{fig:threshold}, we select $\tau = 1.3$ for all datasets.
Moreover, varying threshold $\tau$ values show relatively stable performance of our methods.


\section{More Qualitative Results}
\begin{figure}[h]
    \centering
    \begin{subfigure}{1\linewidth}
        \centering
        \includegraphics[width=1\linewidth]{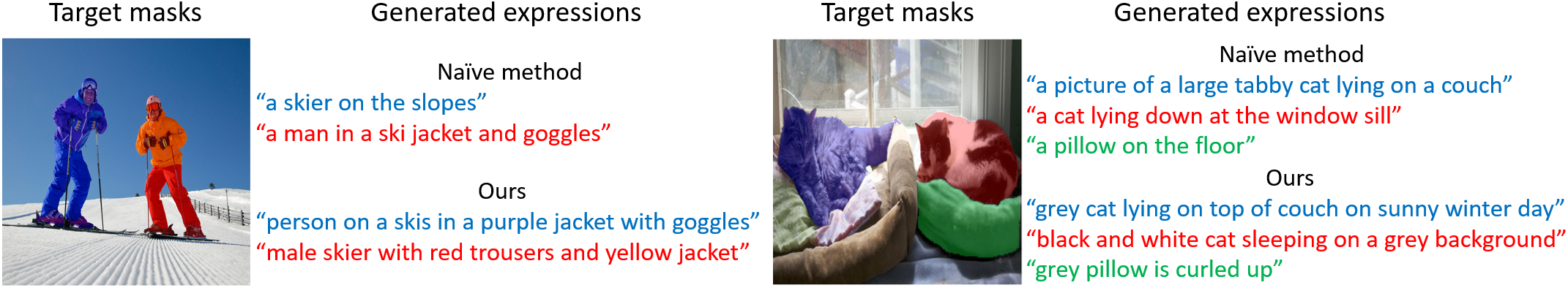}
        \caption{Comparison of our generated expressions with the na\"ive method.}
        \label{sup_fig:qualitative}
    \end{subfigure}
    \hfill
    \begin{subfigure}{1\linewidth}
        \centering
        \includegraphics[width=1\linewidth]{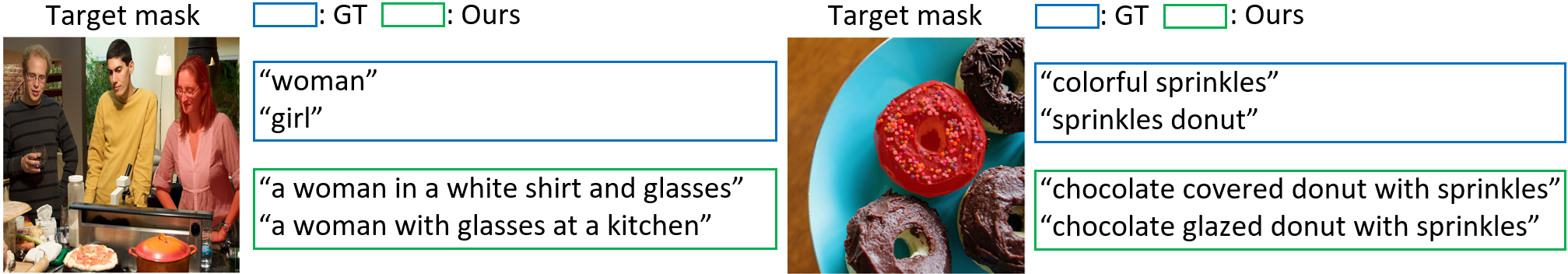}
        \caption{Comparison of our generated expressions with GT expressions.}
        \label{sup_fig:qualitative_with_gt}
    \end{subfigure}
    \caption{Qualitative analysis of our generated expressions compared to the na\"ive methods and GT captions.}
    \label{sup_fig:qual_anylsis}
    \label{sup_fig:qualitative_naylsis}
\end{figure}

We visualize more qualitative results of our generated expressions compared to the na\"ive method and GT expressions in Fig.~\ref{sup_fig:qualitative} and Fig.~\ref{sup_fig:qualitative_with_gt}, respectively.





\section{Semi-supervised Learning Results}
We report the performance in semi-supervised settings compared with recent semi-supervised RIS methods~\cite{semi_ris_1, semi_ris_2, semi_safari} in Table~\ref{tab:semi_results}.
The settings for each method differ. 
We utilize ETRIS as our trained model, while RESMatch and SemiRES employ LAVT as their base model, and SafaRi trains its own models. 
We achieve competitive performance on the RIS benchmark compared to other semi-supervised RIS methods.
\begin{table*}
    \footnotesize
    \centering
    \caption{Semi-supervised comparisons with other semi-ris methods. RESMatch and SemiRES use LAVT~\cite{lavt} as a base RIS model, whereas we adapt ETRIS. SafaRi~\cite{semi_safari} train their proposed framework.}
    \resizebox{\textwidth}{!}{
    \begin{tabular}{c|l|ccc|ccc|ccc}
    \hline
    \multirow{2}{*}{Metric} & \multirow{2}{*}{Method} & \multicolumn{3}{c|}{RefCOCO}  & \multicolumn{3}{c|}{RefCOCO+} & \multicolumn{3}{c}{RefCOCOg} \\ \cline{3-11}
    &       & val   & test A    & test B    & val   & test A    & test B    & val(U)  & test(U) & test(G) \\ \hline
    \multirow{16}{*}{oIoU} &   \multicolumn{1}{c|}{\textbf{0.1\% Training}}  &   &   &   &  &     &   &   &  &  \\ 
    & Supervised~\cite{etris}   & 10.64 & 11.25 & 11.01 & 10.02 & 10.90 & 9.29 & 10.93 & 11.12 & 10.37 \\
    & RESMatch~\cite{semi_ris_2} & 20.07 & 22.47 & 17.88 & 18.11 & 23.54 & 15.37 & 20.97 & 20.91 & - \\
    & Ours     & \textbf{37.89} & \textbf{44.39} & \textbf{30.56} & \textbf{41.15} & \textbf{47.65} & \textbf{33.04} & \textbf{41.50} & \textbf{43.43} & \textbf{43.12} \\ \cdashline{2-11}[1pt/1pt]
    &   \multicolumn{1}{c|}{\textbf{1\% Training}}  &   &   &   &  &     &   &   &  &  \\ 
    & Supervised~\cite{etris}  & 30.12 & 32.85 & 27.86 & 23.34 & 26.33 & 18.83 & 22.72 & 23.30 & 22.97 \\
    & RESMatch~\cite{semi_ris_2} & 36.86 & 42.32 & 31.11 & 31.09 & 35.72 & 26.07 & 32.18 & 32.21 & - \\
    & Ours    & \textbf{46.61} & \textbf{54.27} & \textbf{39.67} & \textbf{43.19} & \textbf{50.54} & \textbf{33.49} & \textbf{43.22} & \textbf{44.40} & \textbf{45.02} \\ \cdashline{2-11}[1pt/1pt]
    &   \multicolumn{1}{c|}{\textbf{5\% Training}}  &   &   &   &  &     &   &   &  &  \\ 
    & Supervised~\cite{etris}  & 39.93 & 45.87 & 33.08  & 36.14 & 41.75 & 28.25 & 30.58 & 32.19 & 32.67 \\
    & RESMatch~\cite{semi_ris_2} & 51.95 & 56.91 & 47.20 & 39.97 & 44.66 & 33.56 & 39.77 & 41.42 & - \\
    & Ours     & \textbf{57.47} & \textbf{63.22} & \textbf{51.46} & \textbf{46.22} & \textbf{53.90} & \textbf{36.31} & \textbf{46.02} & \textbf{47.64} & \textbf{47.73} \\ \cdashline{2-11}[1pt/1pt]
    &   \multicolumn{1}{c|}{\textbf{10\% Training}}  &   &   &   &  &     &   &   &  &  \\ 
    & Supervised~\cite{etris}     & 48.69 & 53.72 & 42.01& 43.18 & 50.72 & 32.87 & 40.65 & 41.64 & 40.73 \\
    & RESMatch~\cite{semi_ris_2} & 58.45 & 62.56 & 54.17 & 45.03 & 51.22 & 37.97 & 45.24 & 47.39 & - \\
    & Ours    & \textbf{61.88} & \textbf{66.96} & \textbf{54.36} & \textbf{49.03} & \textbf{57.02} & \textbf{37.77} & \textbf{48.39} & \textbf{49.66} & \textbf{50.13} \\ \cdashline{2-11}[1pt/1pt]
    \hline \hline
    \multirow{14}{*}{mIoU} &   \multicolumn{1}{c|}{\textbf{0.1\% Training}}  &   &   &   &  &     &   &   &  &  \\ 
    & Supervised~\cite{etris}   & 9.26 & 9.74 & 9.59 & 4.12 & 5.11 & 3.4 & 10.36 & 10.54 & 7.47 \\
    & Ours     & \textbf{42.57} & \textbf{49.70} & \textbf{33.81} & \textbf{46.32} & \textbf{53.32} & \textbf{37.2} & \textbf{45.38} & \textbf{46.14} & \textbf{46.51}  \\ \cdashline{2-11}[1pt/1pt]
    &   \multicolumn{1}{c|}{\textbf{1\% Training}}  &   &   &   &  &     &   &   &  &  \\ 
    & Supervised~\cite{etris}   & 29.65 & 32.44 & 27.64 & 20.69 & 24.33 & 16.48 & 21.65 & 21.88 & 21.01 \\
    & SemiRES~\cite{semi_ris_1}   & \textbf{50.90} & 57.54 & \textbf{44.48} & 36.49 & 42.86 & 28.58 & 34.76 & 36.18 & - \\
    & Ours    & 50.67 & \textbf{57.57} & 43.65 & \textbf{47.93} & \textbf{55.55} & \textbf{37.50} & \textbf{45.95} & \textbf{46.74} & \textbf{47.02} \\ \cdashline{2-11}[1pt/1pt]
    &   \multicolumn{1}{c|}{\textbf{5\% Training}}  &   &   &   &  &     &   &   &  &  \\ 
    & Supervised~\cite{etris} & 45.83 & 53.59 & 35.12  & 36.99 & 43.28 & 28.37 & 30.01 & 30.93 & 30.43 \\
    & SemiRES~\cite{semi_ris_1}   & 61.31 & \textbf{66.64} & \textbf{55.94} & 47.00 & 54.42 & 38.74 & 47.61 & \textbf{50.11} & - \\
    & Ours     & \textbf{61.35} & 65.78 & 55.67 & \textbf{50.58} & \textbf{57.99} & \textbf{39.79} & \textbf{49.18} & 49.59 & \textbf{49.52} \\ \cdashline{2-11}[1pt/1pt]
    &   \multicolumn{1}{c|}{\textbf{10\% Training}}  &   &   &   &  &     &   &   &  &  \\ 
    & Supervised~\cite{etris}   & 50.26 & 55.06 & 43.91& 45.83 & 53.59 & 35.12 & 41.47 & 41.86 & 41.45 \\
    & SafaRi~\cite{semi_safari} & 64.02 & 65.91 & \textbf{61.76} & \textbf{52.98} & 56.24 & \textbf{46.48} & \textbf{52.91} & \textbf{52.94} & - \\
    & Ours    & \textbf{64.09} & \textbf{68.48} & 58.17 & 52.82 & \textbf{60.53} & 42.11 & 50.59 & 51.32 & \textbf{51.33} \\ 
    \hline
    \end{tabular}
    }
    \label{tab:semi_results}
\end{table*}

\end{document}